\documentclass[runningheads]{llncs}

\usepackage[year=2026]{eccv}

\usepackage{eccvabbrv}

\usepackage{graphicx}
\usepackage{booktabs}
\usepackage{xcolor}
\usepackage{amsmath}
\usepackage{algorithm}
\usepackage{algorithmic}
\usepackage{newfloat}
\usepackage{listings}
\usepackage{enumitem}
\usepackage{arydshln}
\usepackage{soul}

\usepackage[accsupp]{axessibility}

\definecolor{cvprblue}{rgb}{0.21,0.49,0.74}
\usepackage{hyperref}

\setlist{nosep,leftmargin=*}

\begin{document}

% \title{AugLift: Uncertainty-Aware Depth Descriptors for Robust 2D-to-3D Pose Lifting}
% \titlerunning{AugLift: Depth Descriptors for 3D Pose Lifting}
\title{AugLift: Depth-Aware Input Reparameterization Improves Domain Generalization in 2D-to-3D Pose Lifting}
\titlerunning{AugLift: Depth-Aware Input Reparameterization for 2D-to-3D Pose Lifting}

\author{Nikolai Warner\inst{1} \and
Wenjin Zhang\inst{2} \and
Hamid Badiozamani\inst{3} \and
Irfan Essa\inst{1} \and
Apaar Sadhwani\inst{3}\thanks{Corresponding author.}}

\authorrunning{N. Warner et al.}

\institute{
Georgia Institute of Technology \and
Rutgers University \and
Amazon \\
\email{\{nwarner30, irfan\}@gatech.edu, wz315@scarletmail.rutgers.edu} \\
\email{badiozam@gmail.com, apaars@gmail.com}
}
\maketitle

%% ========== ABSTRACT ==========
\begin{abstract}
Lifting-based 3D human pose estimation infers 3D joints from 2D keypoints but generalizes poorly because $(x,y)$ coordinates alone are an ill-posed, sparse representation that discards geometric information modern foundation models can recover.
We propose \emph{AugLift}, which changes the representation format of lifting from 2D coordinates to a 6D geometric descriptor via two modules: (1)~an \emph{Uncertainty-Aware Depth Descriptor} (UADD)---a compact tuple $(c, d, d_{\min}, d_{\max})$ extracted from a confidence-scaled neighborhood of an off-the-shelf monocular depth map---and (2)~a scale normalization component that handles train/test distance shifts.
AugLift requires no new sensors, no new data collection, and no architectural changes beyond widening the input layer; because it operates at the representation level, it is composable with any lifting architecture or domain generalization technique.

In the detection setting, AugLift reduces cross-dataset MPJPE by $10.1\%$ on average across four datasets and four lifting architectures while improving in-distribution accuracy by $4.0\%$; post-hoc analysis shows gains concentrate on novel poses and occluded joints.
In the ground-truth 2D setting, combining AugLift with PoseAug's differentiable domain generalization achieves state-of-the-art cross-dataset performance ($62.4$\,mm on 3DHP, $92.6$\,mm on 3DPW; $14.5\%$ and $22.2\%$ over PoseAug), demonstrating that foundation-model depth provides genuine geometric signal complementary to explicit 3D augmentation.
Code will be made publicly available.
\end{abstract}

\section{Introduction}
\label{sec:intro}

\begin{figure}[!ht]
\centering
\includegraphics[width=0.42\textwidth]{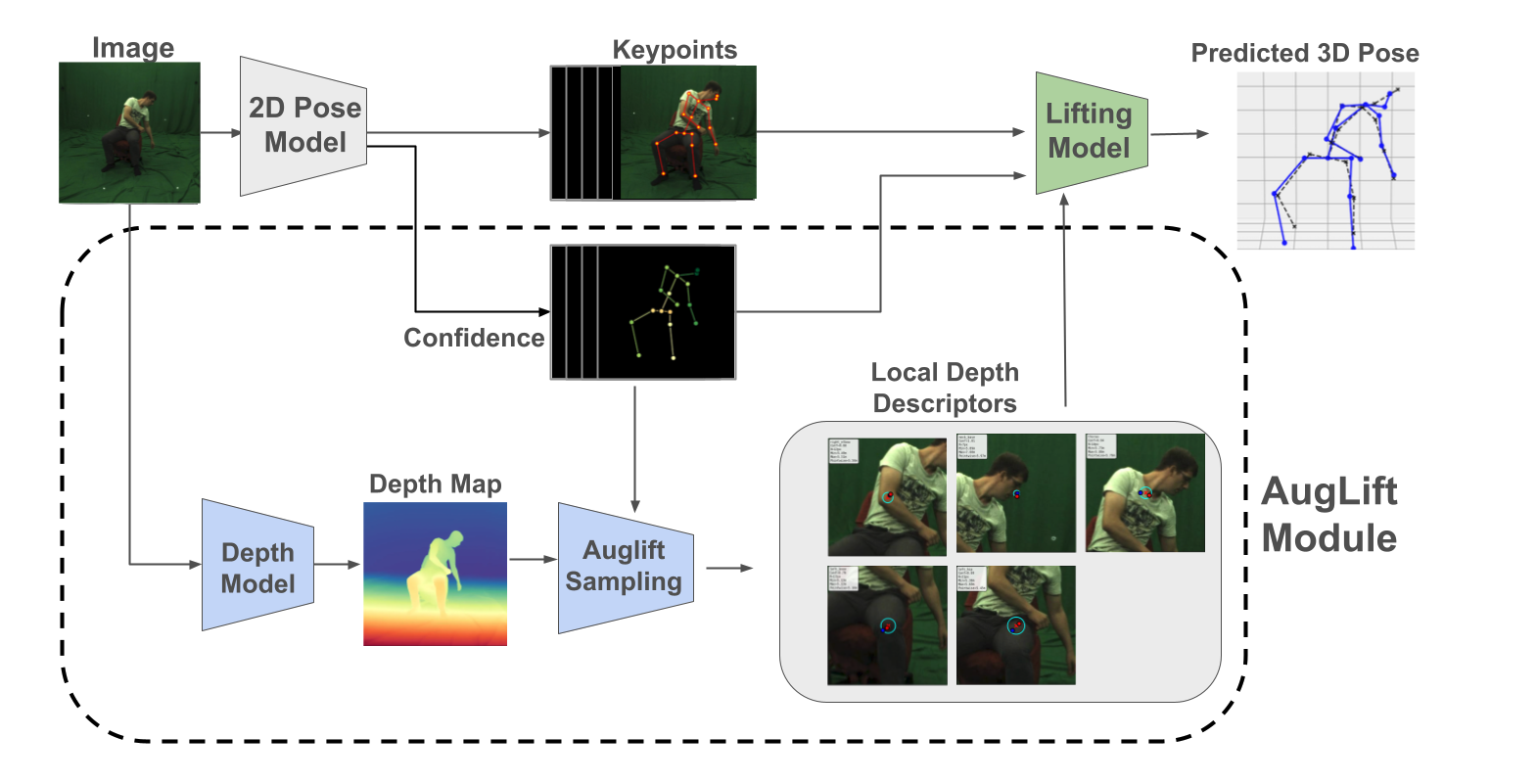}
\caption{\small\textbf{AugLift enriches standard lifting inputs for better generalization.}
Standard lifting uses only sparse 2D coordinates, creating depth ambiguity. AugLift enriches the input with a monocular depth map and keypoint confidence, forming a compact UADD $(c, d, d_{\min}, d_{\max})$ that generalizes to occluded and novel poses.}
\label{fig:system_diagram}
\end{figure}

Estimating 3D human pose from monocular RGB is a long-standing problem in computer vision, with applications in graphics, robotics, AR/VR, and sports analysis.
A widely adopted paradigm decomposes the task into two stages: a 2D keypoint detector first localizes joints in the image, and a \emph{lifting} model then maps these 2D keypoints to 3D joint locations~\cite{martinez2017simple, pavllo2018video, zhu2023motionbert}.
However, lifting models struggle to generalize beyond the lab-style conditions of their training sets, especially in realistic \emph{detection settings} with noisy 2D inputs.
The root cause is that 2D keypoints are an extremely \emph{sparse} representation: they discard geometric cues available in the image that could help disambiguate depth, occlusion, and viewpoint.
Combined with the limited diversity of current training datasets, lifters overfit to the training distribution---state-of-the-art networks achieving $40$--$50$\,mm MPJPE on Human3.6M degrade to over $100$\,mm on in-the-wild datasets such as 3DPW~\cite{manzur2025posebench3d, wang2020predicting}.

A natural response is to enrich the lifting input. Some methods leverage temporal context from video sequences~\cite{pavllo2018video, zhu2023motionbert}, but overfit to motion dynamics and degrade on out-of-distribution actions. Others condition on dense image features~\cite{xu2022monocular, zhou2023lifting}, but risk learning spurious correlations with backgrounds and scene appearance~\cite{zhou2023lifting}.
Meanwhile, modern monocular depth estimators (MDEs) have made striking progress and generalize well across scenes, even though they provide a lower bound for occluded keypoints. MDEs are trained on RGB-D data that requires only a commodity depth sensor to collect, in contrast to the multi-camera MoCap setups needed for 3D pose annotation. Yet their potential as a plug-in signal for 2D--3D lifting remains underexplored.

% This asymmetry creates a scalable improvement lever.
% Multi-camera MoCap to collect new 3D pose data is prohibitively expensive and does not scale, but RGB-D sensors for training MDE models are cheap and abundant---MDE models already generalize across diverse scenes and datasets.
% As MDE models improve with readily available RGB-D data, 3D lifting can improve for free, without additional 3D pose annotation.

In this work we revisit the \emph{input} to lifting and ask: can we improve robustness by augmenting 2D keypoints with noisy, occluded depth cues?
We propose \emph{AugLift}, which consists of two modules: (1)~an \emph{Uncertainty-Aware Depth Descriptor} (UADD)---for each keypoint with detector confidence~$c$, we extract depth statistics from a confidence-scaled neighborhood, forming a compact tuple $(c, d, d_{\min}, d_{\max})$ that captures both local geometry and reliability; and (2)~a \emph{scale normalization} component that handles train/test distance shifts via bounding-box or depth-based rescaling.
AugLift is designed to be \emph{DG-compatible}: its pointwise geometric cues do not depend on scene-level statistics, making it naturally combinable with render-based augmentation methods.
Crucially, AugLift is a change to the \emph{representation format} of lifting---from 2D coordinates to a 6D geometric descriptor---rather than a standalone module. It requires no new sensors, data collection, or architectural changes beyond widening the input layer from $2K$ to $6K$ channels, and is composable with any lifting architecture or domain generalization technique.
Using MDE for lifting is not trivial: MDE estimates are noisy and provide only the nearest visible surface---a lower bound on the true joint depth, not the depth of occluded joints---and exhibit domain shift across datasets. UADD's specific design---confidence-modulated neighborhood size that widens under occlusion, robust summary statistics $(d_{\min}, d, d_{\max})$ that encode this lower bound alongside a central estimate---addresses each of these challenges.

We validate AugLift across three complementary settings.
In the \emph{detection setting}, across four datasets and four architectures, AugLift consistently improves both OOD ($10.1\%$ avg) and ID ($4.0\%$) performance; post-hoc analysis shows gains concentrate on novel poses and occluded joints, where depth statistics resolve front--back ambiguities and confidence regulates sampling neighborhoods.
In the \emph{GT 2D + DG setting}, combining AugLift with PoseAug~\cite{gong2021poseaug} via a novel live depth generation pipeline achieves state-of-the-art cross-dataset results (Section~\ref{sec:poseaug_dg}).
AugLift also complements dense image features, with the compact UADD providing comparable OOD gains to dense learned features while requiring far fewer parameters (Section~\ref{sec:feature_fusion_section}).

\begin{figure}[tb]
    \centering

    % Row 1
    \begin{minipage}{0.14\textwidth}
        \centering
        \includegraphics[width=\linewidth,height=\linewidth]{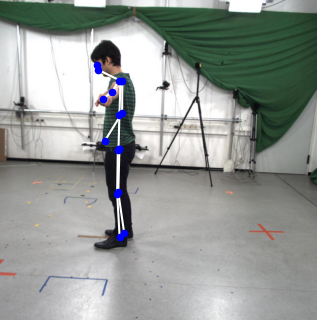}
    \end{minipage}
    \begin{minipage}{0.14\textwidth}
        \centering
        \includegraphics[width=\linewidth,height=\linewidth]{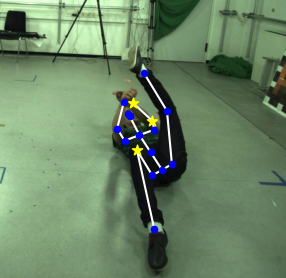}
    \end{minipage}
    \begin{minipage}{0.14\textwidth}
        \centering
        \includegraphics[width=\linewidth,height=\linewidth]{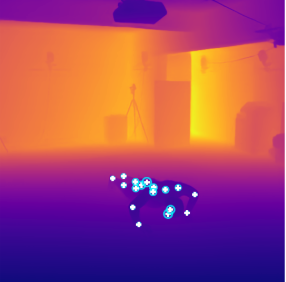}
    \end{minipage}
    \begin{minipage}{0.14\textwidth}
        \centering
        \includegraphics[width=\linewidth,height=\linewidth]{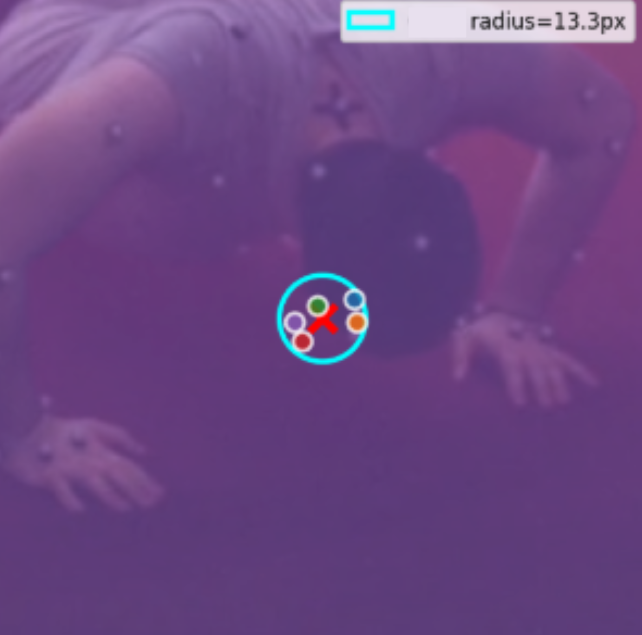}
    \end{minipage}

    % Row 2
    \begin{minipage}{0.14\textwidth}
        \centering
        \includegraphics[width=\linewidth,height=\linewidth]{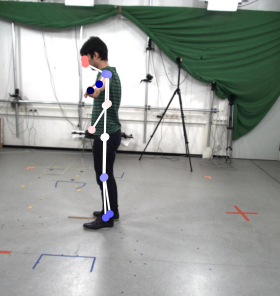}
    \end{minipage}
    \begin{minipage}{0.14\textwidth}
        \centering
        \includegraphics[width=\linewidth,height=\linewidth]{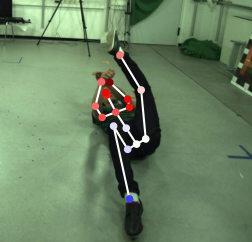}
    \end{minipage}
    \begin{minipage}{0.14\textwidth}
        \centering
        \includegraphics[width=\linewidth,height=\linewidth]{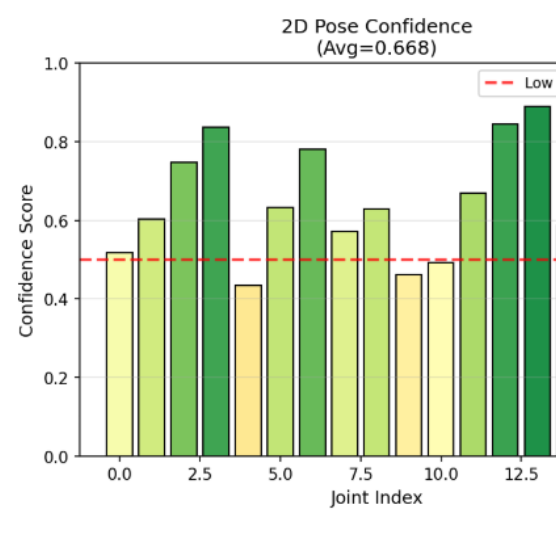}
    \end{minipage}
    \begin{minipage}{0.14\textwidth}
        \centering
        \includegraphics[width=\linewidth,height=\linewidth]{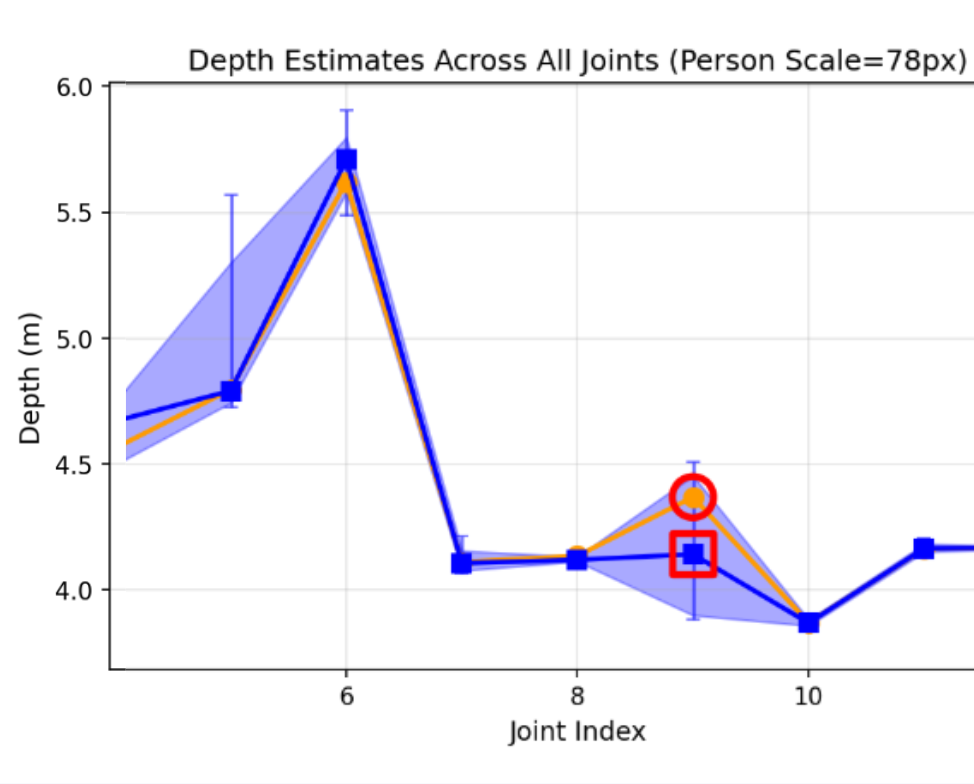}
    \end{minipage}

    \caption{\small\textbf{Confidence-aware depth sampling.} \textbf{Left}: 2D keypoint confidence (top; blue=confident, red\,X=occluded) and monocular depth map (bottom; blue=near, red=far). \textbf{Right}: Low-confidence joints use a wider sampling radius for robust depth statistics; high-confidence joints use a tight radius for precise estimates.}
    \label{fig:depth_confidence}
\end{figure}

\noindent\textbf{Contributions.}
\begin{itemize}
    \item \textbf{C1. AugLift = UADD + Scale Normalization.}
    We propose AugLift, a change to the representation format of 2D$\to$3D lifting from 2D coordinates to a 6D geometric descriptor, consisting of: (1)~UADD, a compact uncertainty-aware depth descriptor from an off-the-shelf MDE, and (2)~scale normalization for train/test distance shifts. Because AugLift operates at the input level, it composes with any lifting architecture or DG technique.

    \item \textbf{C2. SOTA cross-dataset results in GT 2D setting.}
    Combining AugLift with PoseAug~\cite{gong2021poseaug} achieves 62.4\,mm on 3DHP and 92.6\,mm on 3DPW (14.5\% and 22.2\% over PoseAug), demonstrating that foundation model (FM) depth provides genuine geometric signal complementary to explicit 3D augmentation.

    \item \textbf{C3. Systematic detection-setting study.}
    Across 4 datasets $\times$ 4 architectures, AugLift consistently improves performance: 10.1\% OOD and 4.0\% ID MPJPE reductions on average.

    \item \textbf{C4.}
    We provide a detailed post-hoc analysis explaining when and why AugLift works. We show that the largest gains occur on the most challenging cases: novel poses not seen during training (e.g., a 15.7\% error reduction on Fit3D) and significantly occluded joints.
\end{itemize}

\section{Literature Review}
\label{sec:lit_review}

Research in monocular 3D human pose estimation has progressed along several distinct avenues. In the following, we provide a structured review of the relevant literature along these themes—2D-to-3D lifting, generalization strategies, and weakly supervised learning—before discussing methods that enrich the lifting input, which is most relevant to our work.

\noindent\textbf{2D-to-3D Lifting.}
The paradigm of lifting 2D keypoints to 3D space gained prominence after Martinez et al.~\cite{martinez2017simple} demonstrated that a simple, fully-connected network could achieve competitive results by operating directly on 2D coordinates. This approach, however, struggles with the inherent ambiguity of single-frame inputs. To mitigate this, temporal models were introduced. For instance, VideoPose3D~\cite{pavllo2018video} leverages temporal convolutions over sequences of 2D poses to enforce motion consistency. More recently, Transformer-based architectures like MotionBERT~\cite{zhu2023motionbert} have become state-of-the-art by jointly modeling spatial and temporal relationships in human motion. Despite their success, these foundational methods still face challenges with occlusions and depth ambiguities, motivating the need for richer input signals.

\noindent\textbf{Generalization Strategies.}
Improving generalization to unseen datasets and real-world scenarios remains a primary challenge. Several strategies focus on data augmentation and architectural enhancements. PoseAug~\cite{gong2021poseaug}, for example, uses virtual camera augmentations to simulate multiple perspectives during training, thereby improving robustness. Other works propose multitask learning; Wang et al.~\cite{wang2020predicting} integrate 3D pose estimation with camera viewpoint prediction to leverage complementary information. Rhodin et al.~\cite{rhodin2018unsupervised} use a geometry-aware encoder-decoder framework to learn from multi-view images in a semi-supervised fashion. While effective, these methods often introduce significant architectural complexity or require specialized training data.

\noindent\textbf{Weakly-Supervised and Unsupervised Learning.}
Given the high cost of acquiring 3D annotations, methods that reduce reliance on labeled data are crucial. Unsupervised approaches, such as the one by Chen et al.~\cite{chen2019unsupervised}, enforce geometric self-consistency through a lift-reproject-lift cycle, using a 2D pose discriminator to ensure plausible skeletons without any 3D priors. Semi-supervised techniques, explored by Pavllo et al.~\cite{pavllo2018video}, combine limited 3D ground truth with a larger corpus of 2D data by minimizing a reprojection loss. These methods showcase how geometric constraints can substitute for explicit 3D labels.

\noindent\textbf{Enriching Lifting with Image-Derived Cues.}
To overcome the limitations of using only 2D coordinates, another key line of research has explored enriching the input with cues derived directly from the source image. One approach involves creating hybrid models that condition the lifter on rich visual features from a CNN backbone~\cite{nie2019single, xu2022monocular}. However, this strategy faces a critical generalization challenge, as these dense features often learn spurious correlations with the training set's background and appearance~\cite{zhou2023lifting}. To avoid these pitfalls, other works utilize more targeted, semantic signals. For instance, the confidence scores from a 2D detector have been used in isolation to improve robustness to noisy detections~\cite{gong2021poseaug}. Separately, other methods incorporate depth-related information---most commonly as relative ordinal rankings~\cite{pavlakos2018ordinal, jiang2021category}---to address depth ambiguity. We, by contrast, study compact local depth descriptors computed from monocular depth predictions, which improve generalization. As we later show, these sparse descriptors are also highly complementary with dense feature fusion.

%% ========== METHODS ==========

%% ========== METHODS ==========
\section{Methods}
\label{sec:methods}

\subsection{Motivation}
\label{ssec:prelim_analyses}
The design of AugLift was guided by preliminary analyses aimed at understanding the failure modes of modern lifting models (full details in Appendix~\ref{sec:depth_analysis}).

\noindent\textbf{Motion cues can harm generalization.}
While longer motion sequences reduce in-distribution error, we found they typically \emph{degrade} OOD performance (Appendix Fig.~\ref{fig:sequence_length_generalization_TCN}). Models tested on \emph{novel motions composed of familiar poses} (e.g., reversed or sped-up actions) also showed significant drops, indicating overfitting to training motion dynamics rather than static pose geometry.

% \noindent\textbf{Per-frame depth cues offer a robust alternative.}
% Oracle experiments using privileged ground-truth ordinal depth showed that even coarse three-bin depth information reduced cross-dataset error by ${\sim}25\%$ (Appendix Table~\ref{tab:coarse_od_generalization_vp3d}), motivating depth cues as a powerful per-frame signal for generalization.

\noindent\textbf{Per-Frame Depth Cues Offer a Robust Alternative.}
Motivated by this, we investigated the potential of enriching the per-frame input instead.
In contrast to prior works that rely on generic image-derived cues (see Section~\ref{sec:lit_review}), our analysis focused on \textit{depth cues} as a particularly promising signal.
To estimate their upper bound, we conducted oracle experiments using privileged \emph{ground-truth ordinal depth} information at varying granularities.
The results were compelling: augmenting the sparse $(x,y)$ input with even coarse, three-bin ordinal depth information reduced cross-dataset error (H3.6M$\rightarrow$3DPW) by approximately $25\%$ (see Appendix, Table~\ref{tab:coarse_od_generalization_vp3d}).
This suggests that providing the lifter with even basic geometric context via depth cues is a powerful and robust path toward better generalization.

\begin{algorithm}[t]
\caption{The AugLift module.}
\label{alg:auglift_pipeline}
\footnotesize
\textbf{Input:} Image $I$ containing the human subject.

\begin{algorithmic}
    \STATE \textbf{Step 1: Obtain 2D keypoints with confidence scores}
    \STATE \textit{This step is identical to standard lifting.} Run a 2D keypoint detector on $I$ to obtain keypoints with confidence,
    $\{(x_j, y_j, c_j)\}_{j=1}^K$.

    \vspace{3pt}\STATE \textbf{Step 2: Obtain a monocular depth map}
    \STATE Run an off-the-shelf monocular depth estimator on $I$ to obtain a depth map $D$ defined on image pixels.

    \vspace{3pt}\STATE \textbf{Step 3: Compute confidence-aware local depth statistics}
    \FOR{$j = 1, \ldots, K$}
        \STATE \hspace{5pt} i. Convert confidence $c_j \in [0,1]$ to a sampling radius \\
        \hspace{15pt} $r_j = r_{\min} + (1 - c_j)\,(r_{\max} - r_{\min})$.
        \STATE \hspace{5pt} ii. Clamp $r_j$ to $[r_{\min}, r_{\max}]$.
        \STATE \hspace{5pt} iii. Define
        $\mathcal{N}_j = \{(u,v) : \|(u,v) - (x_j,y_j)\|_2 \le r_j\}$.
        \STATE \hspace{5pt} iv. Collect depth values $\mathcal{D}_j = \{D(u,v) \mid (u,v) \in \mathcal{N}_j\}$.
        \STATE \hspace{5pt} v. Compute depth statistics $d_j = \operatorname{median}(\mathcal{D}_j)$, \\
        \hspace{15pt}  $d^{\min}_j = \min(\mathcal{D}_j)$, $d^{\max}_j = \max(\mathcal{D}_j)$.
    \ENDFOR

    \vspace{3pt}\STATE \textbf{Step 4: Rescale 2D keypoints per normalized bounding box}
    \STATE \hspace{5pt} i. Compute keypoint centroid: $c = \bigl(\frac{1}{K}\sum_j x_j, \;\frac{1}{K}\sum_j y_j\bigr)$.
    \STATE \hspace{5pt} ii. Compute box size (average of width and height): \\
        \hspace{15pt} $b = \frac{1}{2}\bigl[(\max_j x_j - \min_j x_j) + (\max_j y_j - \min_j y_j)\bigr]$.
    \STATE \hspace{5pt} iii. Compute scale factor $s =\bar{b}/b$, where $\bar{b}$ is the mean training-set box size.
    \STATE \hspace{5pt} iv. Update $(x_j, y_j) := s \cdot \bigl((x_j, y_j) - c\bigr) + c$ for all $j$.

    \vspace{3pt}\STATE \textbf{Step 5: Normalize confidence and depth statistics}
    \STATE \hspace{5pt} i. Rescale confidence to $[-1,1]$: $\tilde c_j = 2c_j - 1$ for all $j$.
    \STATE \hspace{5pt} ii. Obtain root-relative depths: $\tilde d_j=d_j - d_{\text{root}}$ for all $j$.
    \STATE \hspace{5pt} iii. Clip root-relative depths so $\tilde d_j \in [- \tilde d_{\max}, \tilde d_{\max}]$ for all $j$.
\end{algorithmic}

\textbf{Output:} 6D feature vector
$\tilde q_j = (x_j, y_j, \tilde c_j, \tilde d_j, \tilde d^{\min}_j, \tilde d^{\max}_j)$
for keypoints $j = 1,\ldots,K$. These augmented features serve as inputs to the lifting model.
\end{algorithm}
\subsection{The AugLift Method}
\label{ssec:auglift_method}

AugLift is a lightweight pre-processing pipeline that transforms 2D keypoint inputs into a 6D representation incorporating confidence and local depth statistics, without altering the core lifting architecture (Algorithm~\ref{alg:auglift_pipeline}).

\noindent\textbf{UADD.}
Given keypoints $\{(x_j, y_j, c_j)\}_{j=1}^K$ from a 2D detector and a depth map $D$ from an off-the-shelf MDE, we define a \emph{confidence-aware} radius $r_j = r_{\min} + (1 - c_j)(r_{\max} - r_{\min})$ for each joint. Within the circular neighborhood $\mathcal{N}_j$ of radius $r_j$, we compute three depth statistics---median $d_j$, minimum $d^{\min}_j$, maximum $d^{\max}_j$---forming the UADD: $(c_j, d_j, d^{\min}_j, d^{\max}_j)$.
The key intuition is that $d^{\min}_j$ acts as a geometric lower bound on joint depth (the nearest visible surface), while confidence controls the neighborhood size: tight for visible joints, broad for occluded ones. Full details of each statistic's role are provided in Appendix~\ref{sec:uadd_details}.

\noindent\textbf{Scale normalization.}
AugLift normalizes for train/test distance shifts via two setting-specific instantiations of the same principle.
In the \emph{detection setting}, bounding-box rescaling (Step~4 of Algorithm~\ref{alg:auglift_pipeline}) normalizes 2D skeleton scale by the ratio of box size to the mean training-set box size.
In the \emph{GT+DG setting}, depth-based 2D rescaling (dscale2d, Section~\ref{ssec:auglift_dg}): $(x', y') = (x, y) \cdot d_{\text{root}} / d_{\text{ref}}$ compensates for distance-driven scale shifts.
While DG methods like PoseAug introduce implicit scale invariance through random RT augmentations, AugLift makes this explicit---which explains the large additional gains on 3DPW where the distance domain gap is most pronounced.

\noindent\textbf{Integration.}
After normalizing confidence to $[-1,1]$ and making depth root-relative (Appendix~\ref{sec:uadd_details}), the final per-joint input is
$\tilde q_j = (x_j, y_j, \tilde c_j, \tilde d_j, \tilde d^{\min}_j, \tilde d^{\max}_j)$.
Integration requires only widening the input layer from $2K$ to $6K$ channels; all other layers, losses, and training remain unchanged.

\subsection{Combining AugLift with Domain Generalization}
\label{ssec:auglift_dg}

AugLift's pointwise geometric cues are designed to be combinable with DG and render-based augmentation.
Dense learned image features encode scene-level statistics---texture, lighting, background---which may differ between real images and the synthetic SMPL renders used in augmentation.
By contrast, AugLift's per-joint depth statistics are local and geometric: they depend only on the 3D structure near each joint, not on global scene appearance, enabling them to remain effective even when the depth source shifts from real images to synthetic renders (Section~\ref{sec:poseaug_dg}).

\paragraph{Integration with PoseAug.}
For these experiments we use the XYD input representation (omitting the confidence channel, which is not critical in the GT 2D setting). We integrate PoseAug's differentiable augmentation framework~\cite{gong2021poseaug}, which applies rotation-translation (RT), bone angle (BA), and bone length (BL) transformations in 3D pose space before projection to 2D.  The key challenge is generating geometrically consistent depth for \emph{every augmented pose}---pre-computed depth from original viewpoints cannot be reused after augmentation.

\paragraph{Depth extraction pipeline.}
Naively fitting SMPL~\cite{loper2015smpl} parameters to each augmented skeleton via iterative optimization would make training prohibitively slow.
We develop an analytical BA/BL$\to$SMPL mapping that converts augmented skeletons into SMPL body model parameters. The resulting SMPL meshes are rendered at 128$\times$128 resolution via differentiable rasterization, then processed by Depth Anything V2~\cite{depthanythingv2} to extract monocular depth maps. Per-joint depth values are sampled at 2D keypoint pixel locations via grid sampling, forming the D channel concatenated with detected 2D keypoints (X, Y) as the final XYD input to the lifting network. This pipeline enables live generation of viewpoint- and anatomy-augmented depth signals during training, where the depth channel adapts coherently to each augmented pose rather than using fixed pre-computed depth from original viewpoints. Full details of the SMPL mapping are provided in Appendix~\ref{sec:poseaug_smpl_details}.

\paragraph{Depth-based 2D rescaling (dscale2d).}
The raw XYD representation provides depth as an independent channel but does not account for the relationship between subject distance and 2D projection scale. Subjects closer to the camera appear larger in 2D, creating a systematic domain gap: 3DPW subjects are on average ${\sim}30\%$ closer than Human3.6M training data (mean root depth 3.6m vs.\ 5.15m). We address this with depth-based 2D rescaling, which scales the 2D keypoint coordinates by the ratio of estimated root depth to a reference depth: $(x', y') = (x, y) \cdot d_{\text{root}} / d_{\text{ref}}$. This normalizes 2D scale variations caused by distance differences, allowing the lifting network to see distance-invariant 2D patterns. The reference depth $d_{\text{ref}}$ is a hyperparameter; we find $d_{\text{ref}} = 4.5$m optimal via sweep over the range 3.5--6.0m.

\begin{figure}[tb]
\centering
\includegraphics[trim=0 0 350pt 0, clip, width=0.82\textwidth]{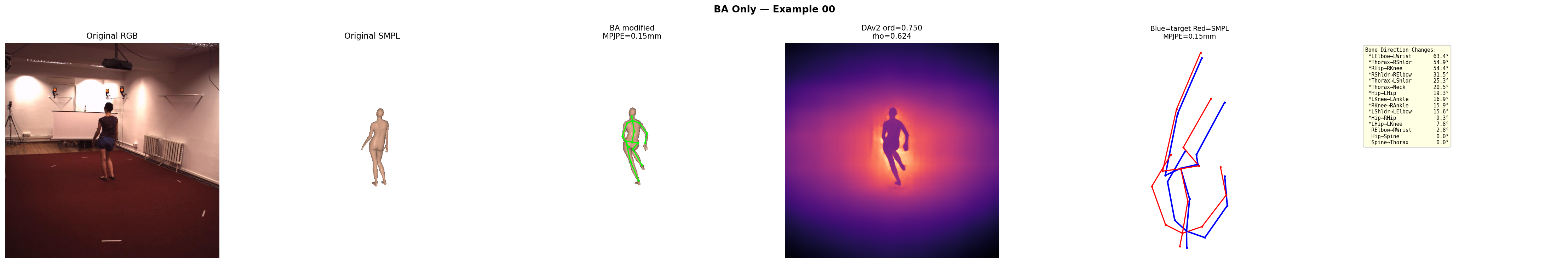}
\caption{\small\textbf{Live depth generation pipeline.} Left to right: (1)~Original RGB frame, (2)~SMPL mesh from original pose, (3)~SMPL mesh after analytical BA augmentation, (4)~Depth Anything V2 depth map from augmented render. Grid sampling extracts per-joint depth to form the D channel, concatenated with 2D coordinates (XY) as input to the lifting network.}
\label{fig:poseaug_pipeline}
\end{figure}

\section{Experiments in the Detection Setting}
\label{sec:experiments}

We evaluate AugLift across three complementary experimental settings. First, we test AugLift in the detection setting across multiple datasets and architectures (this section). Then, we combine AugLift with PoseAug's differentiable domain generalization in the GT 2D setting, demonstrating SOTA cross-dataset performance (Section~\ref{sec:poseaug_dg}). Finally, we compare AugLift's sparse geometric cues with dense learned feature fusion, showing complementarity (Section~\ref{sec:feature_fusion_section}).

\subsection{Experimental Setup}\label{ssec:expt_setup}

\begin{figure}[tb]
    \centering

    \begin{minipage}{0.35\textwidth}
        \centering
        \includegraphics[width=\textwidth]{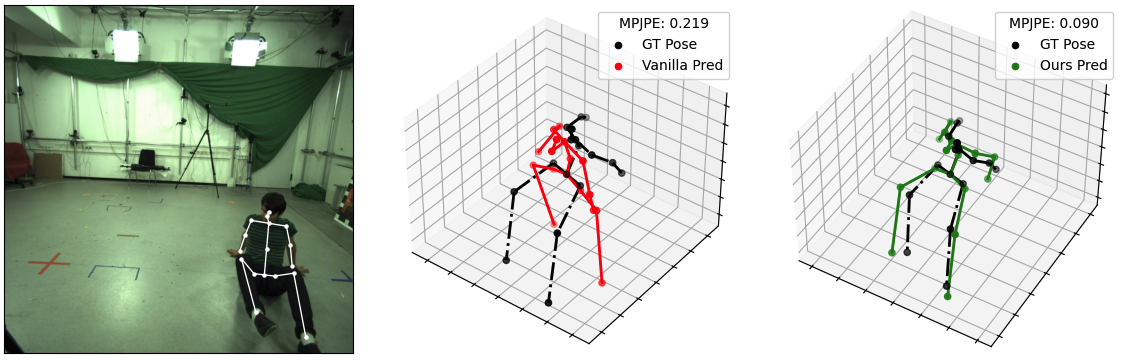}
    \end{minipage}
    \hspace{8mm}
    \begin{minipage}{0.35\textwidth}
        \centering
        \includegraphics[width=\textwidth]{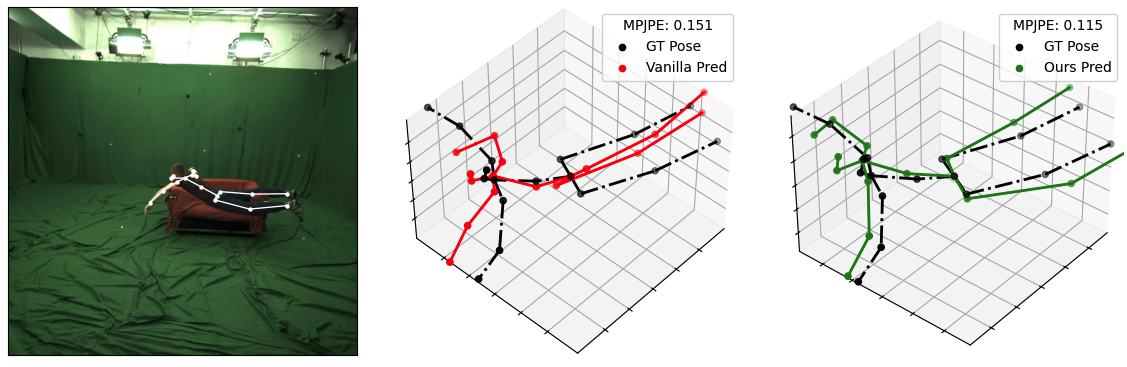}
    \end{minipage}

    \vspace{3mm}

    \begin{minipage}{0.35\textwidth}
        \centering
        \includegraphics[width=\textwidth]{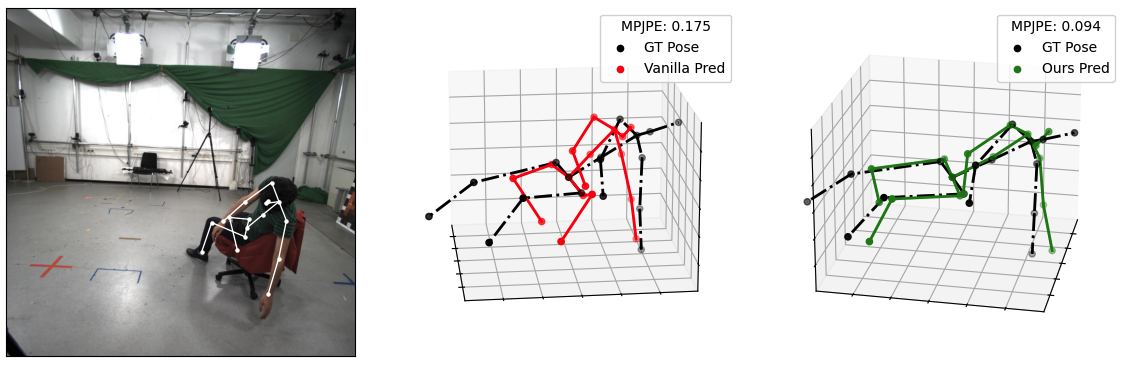}
    \end{minipage}
    \hspace{8mm}
    \begin{minipage}{0.35\textwidth}
        \centering
        \includegraphics[width=\textwidth]{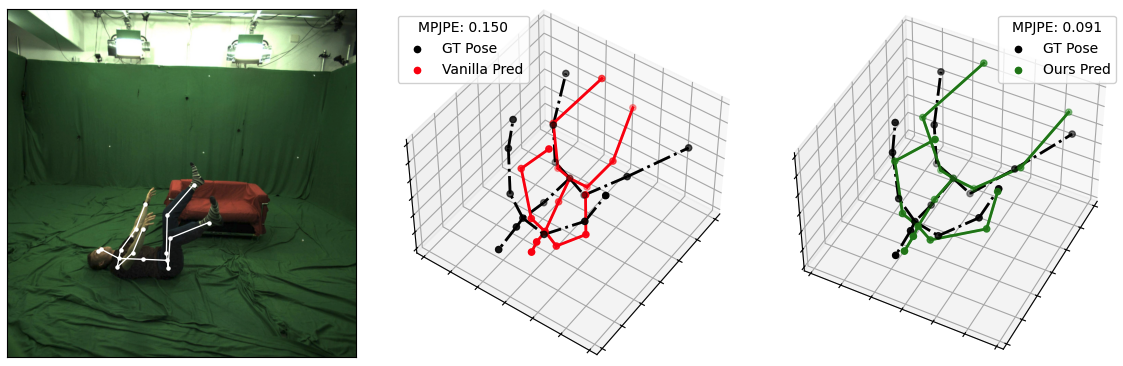}
    \end{minipage}

    \caption{\small\textbf{Qualitative results.} Baseline (red) vs.\ AugLift (green) vs.\ ground truth (black). The baseline fails on OOD poses with occlusion (sitting, crouching); AugLift's depth cues resolve front--back ambiguities.}
    \label{fig:qualitative_results}
\end{figure}

\noindent\textbf{Datasets.}
We conduct experiments on four diverse 3D HPE datasets: Human3.6M (H36M)~\cite{ionescu2014human3}, MPI-INF-3DHP (3DHP)~\cite{mehta2016monocular}, Fit3D~\cite{fieraru2021alfit}, and 3D-Pose-in-the-Wild (3DPW)~\cite{vonmarcard2018recovering}. Each dataset offers unique challenges: H36M consists of controlled indoor motions, 3DHP has greater viewpoint diversity, Fit3D contains varied fitness motions, and 3DPW is a challenging in-the-wild dataset with complex activities and dense occlusions. To thoroughly evaluate generalization, we use H36M, 3DHP, and Fit3D for both training and testing, leveraging their corresponding splits. Due to its smaller size, 3DPW is used exclusively as a test set. 
% While we train on different datasets, H36M serves as our default training set for experiments where the training data is fixed.

\noindent\textbf{Lifting Architectures.}
To demonstrate that AugLift is a modular add-on, we integrate it into four representative lifting backbones: the Transformer-based MotionBERT~\cite{zhu2023motionbert} and PoseFormer~\cite{Zheng2021PoseFormer}, the TCN-based VideoPose3D~\cite{pavllo2018video}, and the MLP-based SimpleBaseline~\cite{martinez2017simple}. Together, these popular models represent a diverse range of lifting paradigms, from simple MLPs and temporal convolutions to state-of-the-art Transformers. As a recent, state-of-the-art model, MotionBERT serves as our default lifting architecture for cross-dataset experiments.

\noindent\textbf{Evaluation Protocol.}
Following standard practice, we measure performance using Mean Per Joint Position Error (MPJPE) in millimeters. We report both in-distribution (ID) performance on the validation split of the training dataset and out-of-distribution (OOD) performance on other datasets to assess generalization. Crucially, our experiments are conducted in the ``detection setting,'' where models are trained and tested on 2D keypoints produced by a detector, not ground truth. This aligns our evaluation with real-world application. 
% To ensure a fair and unified comparison, we use a single off-the-shelf 2D detector (RTMPose-L~\cite{jiang2023rtmpose}) and depth estimation model (Depth Anything v2~\cite{depthanythingv2}) for all experiments. Accordingly, we train both the original backbones and their AugLift-enabled counterparts on the keypoints produced by this detector. As a result, our baseline MPJPE scores may differ slightly from those reported in prior work, which rely on H36M-tuned detectors or ground-truth keypoints that are incompatible with our protocol.

\begin{table}[!t]
\footnotesize
\setlength{\tabcolsep}{3pt}
\renewcommand{\arraystretch}{0.85}
\centering
\caption{\small\textbf{AugLift improves both OOD and ID performance regardless of training dataset.} MPJPE (mm) for MotionBERT baseline vs.\ AugLift across three source datasets. Mean: 10.1\% OOD, 4.0\% ID.}
\begin{tabular}{@{}llrrrr@{}}
\toprule
Train DS & Variant       & H36M   & 3DHP  & Fit3D & 3DPW  \\
\midrule
H36M     & Baseline      & 41.8   & 96.7  & 66.6  & 154.8 \\
         & AugLift       & 40.7   & 85.9  & 55.8  & 146.9 \\
         & $\Delta_{\text{ID}}$  & \textbf{2.6\%} & --          & --          & --          \\
         & $\Delta_{\text{OOD}}$ & --              & \textbf{11.2\%} & \textbf{16.2\%} & \textbf{5.1\%} \\
\midrule
3DHP     & Baseline      & 80.9   & 52.2  & 88.2  & 130.3 \\
         & AugLift       & 79.3   & 50.9  & 76.8  & 121.3 \\
         & $\Delta_{\text{ID}}$  & --              & \textbf{2.5\%}  & --              & --              \\
         & $\Delta_{\text{OOD}}$ & \textbf{2.0\%}  & --              & \textbf{12.9\%} & \textbf{6.9\%}  \\
\midrule
Fit3D    & Baseline      & 127.4  & 166.1 & 39.9  & 195.4 \\
         & AugLift       & 109.1  & 132.9 & 37.1  & 191.3 \\
         & $\Delta_{\text{ID}}$  & --              & --              & \textbf{7.0\%}  & --              \\
         & $\Delta_{\text{OOD}}$ & \textbf{14.4\%} & \textbf{20.0\%} & --              & \textbf{2.1\%}  \\
\bottomrule
\end{tabular}
\label{tab:motionbert_quad}
\end{table}

\subsection{Cross-Dataset Study}

\emph{Does AugLift improve performance on unseen datasets? Does it help—or harm—accuracy on the source dataset itself?} %We conduct a rigorous study to answer these questions.

\noindent\textbf{Setup.} We train a baseline lifting model and its AugLift-enabled counterpart in three independent training runs, each on a different source dataset—H36M, 3DHP, and Fit3D. After training, every model is evaluated on all four datasets (H36M, 3DHP, Fit3D, 3DPW), in line with our Evaluation Protocol in Section~\ref{ssec:expt_setup}. This yields one ID and three OOD scores per run. Using three distinct training distributions ensures our findings are robust and not biased by a single dataset. For this study, we use the state-of-the-art MotionBERT backbone.

\noindent\textbf{Results.} Table~\ref{tab:motionbert_quad} shows that \textbf{AugLift improves accuracy in every ID and OOD scenario}. Averaged across all three training regimes, AugLift improves OOD performance by a mean of \textbf{10.1\%} and also boosts ID performance by a mean of \textbf{4.0\%}. The OOD gains are most pronounced under large domain shifts; for example, training on Fit3D and testing on 3DHP yields a 20.0\% error reduction. These findings demonstrate that AugLift is a powerful mechanism for improving generalization without sacrificing in-domain accuracy.

\subsection{Cross-Architecture Study}\label{ssec:cross_arch_study}

\begin{table}[tb]
\footnotesize
\setlength{\tabcolsep}{2.5pt}
\renewcommand{\arraystretch}{0.85}
\centering
\caption{\small\textbf{Robust gains across diverse architectures.} MPJPE (mm) on H36M (ID) and three OOD sets. Mean: 2.3\% ID, 8.9\% OOD.}
\begin{tabular}{@{}ll rr | rrrr@{}}
\hline
Architecture & Variant & \multicolumn{2}{c|}{ID} & \multicolumn{4}{c}{OOD} \\
\cline{3-8}
 & & H36M & $\Delta$ & 3DHP & Fit3D & 3DPW & $\Delta$ \\
\hline
MotionBERT & Base & 41.8 & --- & 96.7 & 66.6 & 154.8 & --- \\
 & +AugLift & \textbf{40.7} & \textbf{2.6\%} & \textbf{85.9} & \textbf{55.8} & \textbf{146.9} & \textbf{10.8\%} \\
\hline
SimpleBL & Base & 57.5 & --- & 103.8 & 81.6 & 183.7 & --- \\
 & +AugLift & \textbf{54.6} & \textbf{5.0\%} & \textbf{96.1} & \textbf{75.2} & \textbf{147.4} & \textbf{11.7\%} \\
\hline
VP3D & Base & 53.9 & --- & 93.6 & 79.2 & 153.4 & --- \\
 & +AugLift & \textbf{53.8} & \textbf{0.2\%} & \textbf{89.8} & \textbf{75.5} & \textbf{140.8} & \textbf{5.7\%} \\
\hline
PoseFmr & Base & 50.5 & --- & 90.6 & 75.3 & 135.6 & --- \\
 & +AugLift & \textbf{49.8} & \textbf{1.4\%} & \textbf{83.8} & \textbf{70.2} & \textbf{124.5} & \textbf{7.6\%} \\
\hline
\end{tabular}
\label{tab:full_results_aug}
\end{table}

\emph{Do the gains from AugLift span different lifting architectures?} % Having established that AugLift provides robust generalization across datasets, we establish the same across model architectures.

\noindent\textbf{Setup.} We evaluate four distinct and popular lifting backbones: SimpleBaseline, VideoPose3D, PoseFormer, and MotionBERT, as detailed in Section~\ref{ssec:expt_setup}. For each backbone, we train the standard baseline and the full AugLift-enabled model. All models in this study are trained on H36M and evaluated on both in-distribution (H36M) and out-of-distribution (3DHP, Fit3D, and 3DPW) test sets.

\noindent\textbf{Results.} Table~\ref{tab:full_results_aug} confirms that \textbf{AugLift consistently improves performance across all four architectures}. On average, AugLift reduces OOD error by \textbf{8.9\%} and ID error by \textbf{2.3\%}. Notably, the simplest model, SimpleBaseline, receives one of the largest OOD improvements at 11.7\%, suggesting that the explicit geometric cues provided by AugLift may be particularly beneficial for models with less capacity to learn such relationships implicitly.

\subsection{Ablations}
\label{sec:ablations_summary}
\noindent\textbf{Setup.} We conduct ablations along three axes: (i) removing depth and using confidence only, (ii) disabling our per-instance bounding box rescaling, and (iii) varying the temporal context by sweeping the sequence length from 1 to 243 frames. Full results and discussion are provided in Appendix F.

\noindent\textbf{Results.} In summary, using confidence alone yields negligible or inconsistent gains, whereas adding depth on top of confidence produces the strong, consistent ID/OOD improvements reported in Tables~\ref{tab:motionbert_quad} and \ref{tab:full_results_aug}, underscoring the importance of geometric depth cues. Bounding box rescaling delivers substantial improvements on 3DPW (e.g., up to 14.3\% error reduction in Appendix Table~\ref{tab:bbox-rescaling}) and is broadly beneficial even without UADD, making it a generally useful component for handling train–test scale shifts. Finally, the sequence-length study (Appendix Table~\ref{tab:improvements_fixed}) shows that while ID gains taper with longer temporal context, AugLift's OOD gains remain large across all sequence lengths.
% confirming that its per-frame depth cues are complementary to motion and effective in both short- and long-horizon regimes.

\begin{table}[t]
    \footnotesize
    \centering
    \caption{\small\textbf{Gains are largest on novel poses.} Fit3D (trained on H36M) stratified by 3D pose similarity. Baseline fails on novel poses (109.2mm); AugLift reduces by 15.7\%.}
    \label{tab:domain_similarity_fit3d}
    \setlength{\tabcolsep}{3pt}
    \renewcommand{\arraystretch}{0.85}
    \begin{tabular}{@{}lccc@{}}
        \toprule
        Similarity & Base & AugLift & $\Delta$ \\
        \midrule
        Highest 20\% & 55.8 & 53.6 & -3.9\% \\
        Middle 60\% & 67.3 & 66.1 & -1.9\% \\
        Lowest 20\% & 109.2 & 92.1 & -15.7\% \\
        \bottomrule
    \end{tabular}
\end{table}

\subsection{When and Why Does AugLift Help?}
\label{ssec:when_why}

\noindent\textbf{Setup.}
To understand where AugLift's gains come from, we perform a post-hoc analysis using a PoseFormer model (sequence length 27) trained on H36M and evaluated with and without AugLift on three OOD test sets: Fit3D, 3DHP, and 3DPW.
We design three analyses, each targeting a specific challenge in 3D pose estimation.
In each case, we bucket test frames using an oracle metric and compare the baseline and AugLift within each bucket:
\begin{itemize}
    \item \emph{Occlusion status.} To measure the impact of self-occlusion, we stratify test frames into quartiles (Q1–clear to Q4–most occluded) using a geometric occlusion proxy derived from the ground-truth 3D pose, which detects 2D keypoint overlap based on 3D depth ordering.
    \item \emph{Pose similarity.} To quantify how AugLift behaves on poses not seen in training, we cluster H36M training poses into a set of 3D prototypes and, for each test frame, compute the distance to the nearest prototype. This defines three buckets: "Highest 20\%", "Middle 60\%", and "Lowest 20\%" similarity (Table~\ref{tab:domain_similarity_fit3d}).
    \item \emph{2D keypoint reliability.} To probe robustness to noisy inputs, we measure mean 2D error (pixels) between detected and ground-truth keypoints and stratify frames into quartiles from "Q1–Best" (lowest 2D error) to "Q4–Worst" (highest 2D error).
\end{itemize}
We complement these quantitative analyses with qualitative examples in Figure~\ref{fig:qualitative_results} on challenging poses such as sitting and crouching.

\noindent\textbf{Findings.}
We observe three consistent patterns:
\begin{itemize}
    \item \emph{Tail-error reduction.} AugLift primarily reduces worst-case failures: on 3DHP, p90 error drops by 11.1\% (154.1\,mm $\rightarrow$ 136.9\,mm), while the median (p50) improves by only 3.5\% (80.6\,mm $\rightarrow$ 77.8\,mm), with a similar pattern on 3DPW.
    \item \emph{Occlusion sensitivity.} Gains grow with occlusion. On Fit3D, the least occluded quartile sees a 4.1\% median improvement (64.5\,mm $\rightarrow$ 61.9\,mm), versus 9.4\% (79.6\,mm $\rightarrow$ 72.1\,mm) in the most occluded quartile.
    \item \emph{Pose novelty.} AugLift helps most on poses farthest from the H36M training distribution. On Fit3D (Table~\ref{tab:domain_similarity_fit3d}), the "Lowest 20\%" similarity bucket achieves a 15.7\% MPJPE reduction (109.2\,mm $\rightarrow$ 92.1\,mm), compared to 3.9\% and 1.9\% in the highest and middle buckets.
\end{itemize}

\noindent\textbf{Mechanism: confidence–depth synergy.}
These trends, together with the qualitative examples in Figure~\ref{fig:qualitative_results}, support a simple mechanism view.
When 2D confidence is very low and keypoints are badly misplaced, both the baseline and AugLift can fail, since depth is sampled around the wrong location.
For \emph{moderate} confidence, typically corresponding to partially occluded joints, AugLift provides its largest gains: the $d_{\min}$ statistic acts as a lower bound on true joint depth, helping avoid front--back flips even when the keypoint location is uncertain.
Finally, AugLift can help even for \emph{high}-confidence joints, where 2D locations are accurate but depth remains ambiguous (e.g., bending forward vs.\ backward). Here, the median depth $d$ closely tracks the true joint depth and resolves this ambiguity.
In all regimes, confidence controls the neighborhood size and determines whether depth should be treated as a precise estimate or a conservative bound, making AugLift especially beneficial on occluded, novel, and depth-ambiguous poses.

\section{AugLift with Domain Generalization}
\label{sec:poseaug_dg}

Having established AugLift's effectiveness in the detection setting (Section~\ref{sec:experiments}), we now evaluate it in a complementary regime: the ground-truth 2D keypoint setting with explicit domain generalization.
This tests whether AugLift's depth signal is additive with 3D geometric augmentation---a stronger test of whether FM depth provides genuine geometry beyond what augmentation alone can achieve.
We combine AugLift (XYD + dscale2d) with PoseAug~\cite{gong2021poseaug}, training on H36M only with GT 2D keypoints. The live depth generation pipeline (Section~\ref{ssec:auglift_dg}) provides geometrically consistent depth for every augmented pose.
We compare against published DG methods: PoseAug, DH-AUG~\cite{huang2022dhaug}, CEE-Net~\cite{li2023ceenet}, PoseGU~\cite{guan2023posegu}, and DAF-DG~\cite{peng2024daf}. We include a comprehensive reproduction of DAF-DG showing its published SOTA is unreproducible from official code (details in Appendix~\ref{sec:poseaug_smpl_details}).

\begin{table}[tb]
\footnotesize
\setlength{\tabcolsep}{3pt}
\renewcommand{\arraystretch}{0.85}
\centering
\caption{\small\textbf{Cross-dataset evaluation (MPJPE~$\downarrow$, mm).} Train on H36M only. $^\dagger$DAF-DG~\cite{peng2024daf} unreproducible; best repro = 80.5mm.}
\label{tab:poseaug_crossdataset}
\begin{tabular}{@{}lcc@{}}
\toprule
Method & 3DHP & 3DPW \\
\midrule
VPose~\cite{pavllo2018video} & 102.3 & 125.7 \\
PoseGU~\cite{guan2023posegu} & 75.0 & --- \\
PoseAug~\cite{gong2021poseaug} & 73.0 & 119.0 \\
DH-AUG~\cite{huang2022dhaug} & 71.2 & 112.8 \\
CEE-Net~\cite{li2023ceenet} & 69.7 & --- \\
DAF-DG~\cite{peng2024daf}$^\dagger$ & 63.1 & 106.6 \\
\midrule
AugLift + PoseAug (XYD) & 62.9 & 104.3 \\
AugLift + PoseAug (XYD+dscale2d) & \textbf{62.4} & \textbf{92.6} \\
\bottomrule
\end{tabular}
\end{table}

\subsubsection{Results.}
\label{ssec:dafdg_comparison}

As shown in Table~\ref{tab:poseaug_crossdataset}, AugLift+PoseAug (simply AugLift combined with PoseAug) achieves state-of-the-art cross-dataset performance on both benchmarks. On 3DHP test, our XYD+dscale2d model achieves 62.4mm MPJPE, a 14.5\% relative improvement over PoseAug~\cite{gong2021poseaug} (73.0mm) and 8.8mm below the next reproducible published result (DH-AUG~\cite{huang2022dhaug}, 71.2mm). On 3DPW, we achieve 92.6mm, a 22.2\% relative improvement over PoseAug (119.0mm) and 20.2mm below DH-AUG (112.8mm). The previously claimed SOTA (DAF-DG~\cite{peng2024daf}: 63.1mm 3DHP, 106.6mm 3DPW) is unreproducible from official code---our best reproduction achieves 80.5mm.

\subsubsection{Key Findings.}
XYD alone already outperforms all published methods (62.9\,mm 3DHP); adding dscale2d provides a further 0.5\,mm on 3DHP and 11.7\,mm on 3DPW, confirming that both depth and scale normalization contribute, with especially large gains on 3DPW where the distance domain gap is most pronounced.
Notably, AugLift's depth signal is trained on SMPL renders that are domain-shifted from real images, yet still provides large gains---validating that sparse geometric cues are robust to the render-vs-real gap that would break dense learned features.

% \begin{enumerate}[nosep,leftmargin=*]
%     \item \textbf{Depth augmentation helps}: Adding monocular depth and depth-based 2D rescaling reduces MPJPE from 74.0mm (PoseAug) to 62.4mm on 3DHP (15.7\% reduction) and from 119.0mm to 92.6mm on 3DPW (22.2\% reduction).
%     \item \textbf{Complementary gains}: XYD without rescaling already outperforms all published methods (62.9mm 3DHP); adding dscale2d provides a further 0.5mm improvement on 3DHP and 11.7mm on 3DPW, confirming that both depth and scale contribute.
%     \item \textbf{Cross-dataset generalization}: The combination of depth-augmented input and PoseAug's differentiable augmentation framework improves robustness to domain shift across both benchmarks, with especially large gains on 3DPW where the distance domain gap is most pronounced.
%     \item \textbf{DG-compatibility confirmed}: AugLift's depth signal, trained on SMPL renders that are domain-shifted from real images, still provides large gains---validating that sparse geometric cues are robust to the render-vs-real domain gap that would break dense learned features.
% \end{enumerate}

%% ========== DENSE FEATURES ==========
\section{Dense Features and AugLiftV2}
\label{sec:feature_fusion_section}

Sections~\ref{sec:experiments}--\ref{sec:poseaug_dg} established that AugLift's sparse, interpretable UADD descriptor provides robust gains across both detection and DG settings.
A natural question is whether \emph{denser} features---either from the 2D detector (image features) or the depth backbone (depth-image features)---can provide additional gains.
We find that dense features are powerful for in-distribution accuracy but carry a generalization risk: they can overfit to domain-specific texture and appearance.
By contrast, AugLift's sparse geometric cues function as a stable \emph{geometry bottleneck}, providing consistent gains even under large domain shifts.
The two signal types are complementary.

\begin{figure}[!t]
    \centering
    \includegraphics[width=0.85\linewidth]{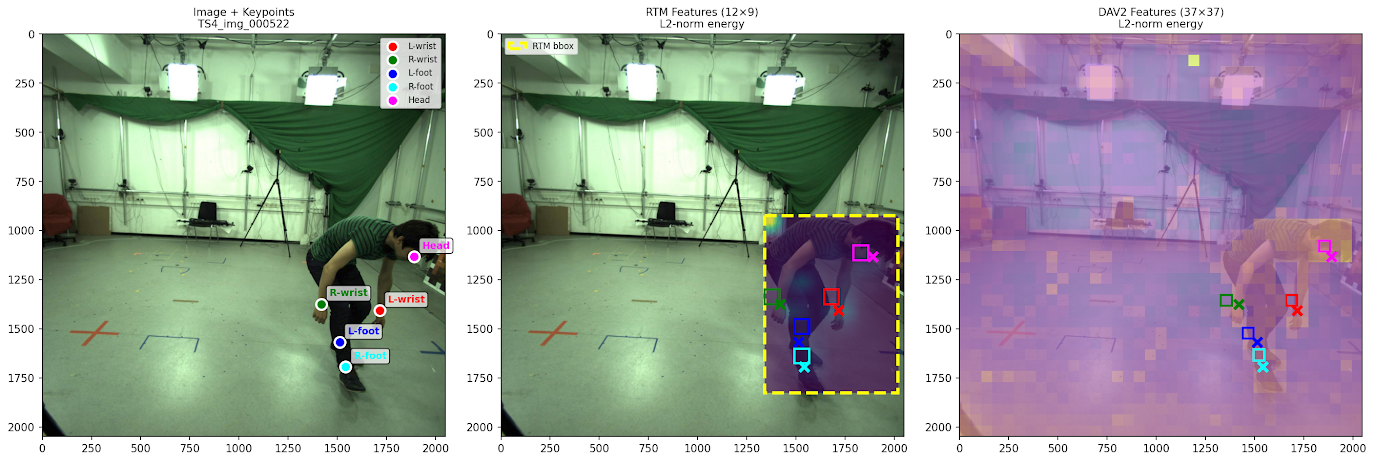}
    \caption{\small\textbf{Sampling dense feature maps at 2D keypoints.}
    We show the 2D detection bounding box from RTMPose (center) aligned with its feature map, alongside the monocular depth feature map (right). We visualize the $\ell_2$ norm of channel activations per coarse patch and sample features at 2D keypoints by averaging the four nearest spatial embeddings. This design preserves the lightweight nature of standard lifting and AugLift while enabling the use of richer spatial cues from image and depth features.}
    \label{fig:feature_fusion}
\end{figure}

\begin{table}[t]
    \centering
    \footnotesize
    \caption{\small\textbf{Sparse cues complement dense features.}
    PoseFormer (slen=1), H36M train. RTMPose = image features; AugLiftV2 = depth-image features.}
    \label{tab:feature_map_fusion_combined}
    \setlength{\tabcolsep}{3pt}
    \renewcommand{\arraystretch}{0.85}
    \begin{tabular}{@{}lcccc@{}}
    \toprule
    Config & Val & $\Delta_{\text{ID}}$ & 3DHP & $\Delta_{\text{OOD}}$ \\
    \midrule
    Base Model          & 54.3 & --    & 94.5 & --    \\
    Base Model + RTMPose& 51.1 & +5.9 & 88.1 & +6.8 \\
    \midrule
    AugLift             & 51.7 & +4.8 & 87.9 & +7.0 \\
    AugLift + RTMPose   & 50.3 & +7.4 & 84.6 & +10.5 \\
    \midrule
    AugLiftV2           & 47.9 & +11.8 & 87.1 & +7.8 \\
    AugLiftV2 + RTMPose & 47.9 & +11.8 & 85.9 & +9.1 \\
    \bottomrule
    \end{tabular}
\end{table}

\subsection{Complementarity with Image Features}
\label{ssec:image_features}

AugLift uses sparse, interpretable cues (UADD) attached to 2D keypoints.
In parallel, prior work has shown that fusing dense image features from the 2D detector into the lifter can improve accuracy, especially in the single-frame setting (Lifting-by-Image~\cite{zhou2023lifting}, CAPoseFormer~\cite{zhao2023caposeformer}).
Here we ask: \emph{are these two approaches complementary?}

\noindent\textbf{Setup.} We follow the detection setting from Section~\ref{ssec:expt_setup} and train on H36M.
We mimic a single-frame CAPoseFormer to utilize image features from RTMPose, both with and without AugLift.
From RTMPose, we extract a coarse feature map and sample features at each detected keypoint location by averaging the four nearest spatial patches (Figure~\ref{fig:feature_fusion}).
The sampled features are linearly projected and concatenated with the keypoint tokens before the transformer encoder. Other details, including feature map sizes, are provided in Appendix G.

\noindent\textbf{Results.}  Table~\ref{tab:feature_map_fusion_combined} shows three main trends.
First, adding image features alone improves both ID and OOD performance over the base model (e.g., $+5.9\%$ ID and $+6.8\%$ OOD).
Second, AugLift by itself provides comparable gains ($+4.8\%$ ID, $+7.0\%$ OOD).
Third—and most importantly—the combination \textbf{AugLift + RTMPose} is best overall, improving ID by $7.4\%$ and OOD by $10.5\%$.
This establishes that AugLift's sparse, confidence-aware depth cues and dense image features carry \emph{complementary} information, and that using both yields the strongest cross-dataset performance.
However, dense image features are tied to the 2D detector's learned representations, which can overfit to scene-specific statistics that shift across datasets; AugLift's sparse depth cues, by contrast, remain effective even under large domain shifts (Sections~\ref{sec:experiments}--\ref{sec:poseaug_dg}) and when applied to synthetic SMPL renders (Section~\ref{sec:poseaug_dg}), making it a more reliable foundation for cross-dataset generalization.

\subsection{AugLiftV2: Learning Depth-Image Features}
\label{ssec:depth_features}

Motivated by the success of image-feature fusion, we next explore whether similar dense features extracted from the \emph{monocular depth image} can further strengthen AugLift.
We define \emph{AugLiftV2} as an extension of AugLift that additionally uses these depth-image features.

\noindent\textbf{Setup.} In AugLiftV2, we sample depth-backbone feature maps at 2D keypoints using the same strategy as before (see Figure~\ref{fig:feature_fusion}) and pass them through a small encoder before concatenation with the keypoint tokens.
Unlike UADD, these depth-image features are learned and no longer directly interpretable as simple statistics $(c, d, d_{\min}, d_{\max})$. Architectural details are deferred to Appendix G.

\noindent\textbf{Results.} As shown in  Table~\ref{tab:feature_map_fusion_combined}, AugLiftV2 yields a stronger ID improvement than AugLift (Val 51.7\,mm $\rightarrow$ 47.9\,mm; $+11.8\%$ vs.\ $+4.8\%$) and slightly better OOD performance (87.9\,mm $\rightarrow$ 87.1\,mm; $+7.8\%$ vs.\ $+7.0\%$).
Moreover, AugLiftV2 remains complementary to image features: combining AugLiftV2 with RTMPose improves OOD error from 88.1\,mm to 85.9\,mm ($+9.1\%$ over the base model), while preserving the strong ID gains.
The larger ID gain but marginal OOD improvement of AugLiftV2 over UADD is consistent with the general pattern: richer learned features improve in-domain accuracy but provide diminishing returns for cross-dataset generalization, where the interpretable, compact UADD descriptor proves more robust.

\section{Conclusion}
\label{sec:conclusion}

We introduced AugLift, a change to the representation format of 2D$\to$3D lifting consisting of two modules: an Uncertainty-Aware Depth Descriptor (UADD) and a scale normalization component.
In the detection setting, AugLift reduces cross-dataset error by 10.1\% and in-distribution error by 4.0\% across four datasets and four lifting architectures, with gains concentrating on occluded joints and novel poses.
In the GT 2D setting, combining AugLift with PoseAug achieves state-of-the-art cross-dataset performance (62.4\,mm on 3DHP, 92.6\,mm on 3DPW; 14.5\% and 22.2\% over PoseAug).
AugLift's sparse geometric cues complement dense image features but, crucially, remain stable under domain shift and synthetic rendering---functioning as a geometry bottleneck for robust lifting.
Together, these results establish monocular depth as a scalable improvement lever for 3D pose lifting: as MDE models improve with readily available RGB-D data, lifting improves without additional 3D pose annotation.

Looking ahead, the depth estimator could be distilled into a lighter-weight unified architecture for real-time deployment.
Our findings on the brittleness of motion priors highlight the need to understand how much temporal context remains necessary when strong per-frame cues like UADD are available.
Finally, AugLift-style descriptors could be extended to multi-person scenes, other 3D prediction tasks, and settings with weaker or self-supervised depth signals.

\newpage
\clearpage

{
    \small
    \bibliographystyle{splncs04}
    \bibliography{main}
}
\clearpage
% Appendices built separately — see appendices.tex
% Appendices — included via \input{appendices} from main_full.tex
% No preamble: all packages and commands are inherited from the main document.

\appendix

\section*{Appendix: UADD Design Details}
\label{sec:uadd_details}

This appendix provides additional details on the UADD descriptor and channel normalization, supplementing Section~3.2 of the main paper.

\paragraph{Role of each depth statistic.}
The three summary statistics in $\text{UADD}_j = (c_j, d_j, d^{\min}_j, d^{\max}_j)$ each serve a distinct purpose.
The minimum $d^{\min}_j$ captures the depth of the nearest visible surface in the sampling neighborhood and acts as a geometric lower bound on the true joint depth---particularly informative under occlusion, where the depth map reflects the occluder rather than the joint itself.
The maximum $d^{\max}_j$ reflects more distant surfaces and flags potential background contamination when the sampling neighborhood extends beyond the subject.
The median $d_j$ provides a robust central estimate that is close to the true joint depth when the joint is visible, and degrades gracefully under partial occlusion.
Together, these three statistics turn a single noisy depth value into a compact descriptor that encodes local 3D structure and its reliability.

\paragraph{Confidence-aware neighborhood.}
The confidence $c_j$ from the 2D detector controls the sampling radius via $r_j = r_{\min} + (1 - c_j)(r_{\max} - r_{\min})$, clamped to $[r_{\min}, r_{\max}]$.
For high-confidence (visible) joints, the small radius yields precise, localized depth statistics.
For low-confidence (often occluded) joints, the expanded radius gathers depth over a broader region, providing a more conservative signal that acknowledges keypoint localization uncertainty.
In our experiments we use $r_{\min}=3$ and $r_{\max}=30$ pixels.

\paragraph{Channel normalization.}
Before passing to the lifting model, confidence is mapped to $[-1,1]$ via $\tilde c_j = 2c_j - 1$, which centers the signal and provides a natural zero-information default when dropout zeroes out the confidence channel during training.
Depth statistics are made root-relative ($\tilde d_j = d_j - d_{\text{root}}$) and clipped to a maximum absolute value of $\tilde d_{\max} = 2$\,m.
In the detection setting, models are trained and evaluated on detected keypoints, which is necessary to obtain meaningful confidence scores.

\section*{Appendix A: Experimental Setup Details}
\label{sec:lifting_prelims}

This appendix provides additional details on our experimental setup, supplementing Section~4.1 of the main paper.

\paragraph{Prediction Tasks.}
The lifting literature considers two distinct settings: using ground-truth 2D keypoints ("GT setting") and using detected 2D keypoints ("detection setting"). In the GT setting, 2D keypoints are obtained directly from dataset annotations, which eliminates noise and allows for controlled studies that isolate specific effects (e.g., the impact of sequence length). In the detection setting, keypoints are extracted from a 2D pose detector, which introduces real-world noise. Unless stated otherwise for a specific preliminary analysis, all main experiments in this paper are conducted in the detection setting to best align with practical applications.

\paragraph{Dataset Descriptions.}
Our experiments rely on four large-scale datasets. Human3.6M (H36M)~\cite{ionescu2014human3} is a widely used indoor dataset featuring controlled motion sequences and fixed cameras; while densely annotated, its limited diversity can hinder generalization. In contrast, MPI-INF-3DHP (3DHP)~\cite{mehta2016monocular} introduces greater viewpoint diversity with more varied camera angles and occlusions. Fit3D~\cite{fieraru2021alfit} focuses on fitness-related motions with significant self-occlusions. Finally, 3D-Pose-in-the-Wild (3DPW)~\cite{vonmarcard2018recovering} was captured "in the wild" and includes complex activities like skateboarding, making it one of the most challenging benchmarks for real-world generalization.

\paragraph{Evaluation Details for 3DPW and Fit3D.}
To ensure fair comparisons, we made two dataset-specific adjustments. First, for 3DPW, we standardize the ground-truth skeletons to match the proportions of the H36M format by recentering the torso and proportionally adjusting the limbs. While performance is typically benchmarked on the official 3DPW test set, we evaluate on the full dataset, as it is used exclusively for OOD testing in our protocol and this provides a more comprehensive assessment. Second, for cross-dataset experiments requiring training on Fit3D, we create our own splits from the publicly available subjects (training on s03-s09, validating on s10-s11), as its official test set is held-out.

\paragraph{Implementation Note on PoseFormer.}
For the PoseFormer architecture, we report results using a sequence length of 27 frames. We successfully reproduced the 9-frame and 27-frame models from the original work to within a few millimeters of the reported MPJPE. However, despite extensive hyperparameter tuning, we were unable to replicate the results for the 81-frame model, a challenge that has been noted by other researchers in public forums.

\section*{Appendix B: AugLift Performance by Architecture}

This section provides a high-level summary of AugLift's performance across the four tested architectures, complementing the detailed results in the main text. The values in Table~\ref{tab:arch_gains} are obtained by averaging the percentage improvements across the various input sequence lengths relevant to each model. Specifically, the sequence lengths averaged are: 1 frame for SimpleBaseline~\cite{martinez2017simple}; 9, 27, and 81 frames for PoseFormer~\cite{Zheng2021PoseFormer}; and 9, 27, and 243 frames for both MotionBERT~\cite{zhu2023motionbert} and VideoPose3D~\cite{pavllo2018video}.

\begin{table}
\centering
\caption{\textbf{Average performance gains from AugLift across different model architectures.} The table shows the percentage reduction in MPJPE, averaged over multiple sequence lengths for each backbone. Gains are shown relative to a standard baseline ($\Delta_{\text{AugLift}}$) and a baseline that already includes 2D confidence scores ($\Delta_{\text{Depth}}$).}
\begin{tabular}{l l r r}
\toprule
Architecture    & Split & $\Delta_{\text{AugLift}}$ & $\Delta_{\text{Depth}}$ \\
\midrule
SimpleBaseline & ID    &  5.0\% &  6.7\% \\
               & OOD   &  11.7\% & 9.2\% \\
\midrule
PoseFormer      & ID    &  1.1\% &  2.1\% \\
                & OOD   &  6.2\% &  5.6\% \\
\midrule
MotionBERT      & ID    &  3.7\% &  1.6\% \\
                & OOD   & 10.2\% &  5.7\% \\
\midrule
VideoPose3D     & ID    &  2.7\% &  2.6\% \\
                & OOD   &  5.9\% &  4.9\% \\
\bottomrule
\end{tabular}
\label{tab:arch_gains}
\end{table}

\section*{Appendix C: Additional Ablation Studies}
\label{sec:ablations_full}

To validate our design choices and explore training strategies, we performed several ablation studies. This included a hyperparameter search for the SimpleBaseline model and an investigation into the effect of our bounding box rescaling method at inference time. The results are presented in Table~\ref{tab:sbs_ablation} (hyperparameter search) and Table~\ref{tab:bbox_rescale_results} (bounding box rescaling analysis).

\begin{table}
  \centering
  \small
  \caption{\textbf{Hyperparameter ablation for the SimpleBaseline + AugLift model, trained on H36M.} The top table shows the effect of varying dropout (with network depth fixed at 2 blocks), while the bottom shows the effect of changing network depth (with dropout fixed at 0.5). The batch size was 64 for all runs.}
  \label{tab:sbs_ablation}
  \begin{tabular}{@{}l r r r r@{}}
    \toprule
    \multicolumn{5}{c}{\textbf{Dropout Ablation (blocks = 2)}} \\
    \midrule
    Dropout & H36M & 3DHP & Fit3D & 3DPW \\
    \midrule
    0.25    & 54.6 & 97.1  & 72.2  & 163.2 \\
    0.33    & 54.7 & 96.6  & 72.5  & 163.6 \\
    0.42    & 54.9 & 96.0  & 73.9  & 164.4 \\
    0.50    & 56.0 & 96.2  & 77.2  & 167.8 \\
    \midrule
    \multicolumn{5}{c}{\textbf{Block-Depth Ablation (dropout = 0.5)}} \\
    \midrule
    Blocks  & H36M & 3DHP  & Fit3D & 3DPW \\
    \midrule
    2       & 55.6 & 99.6  & 75.1  & 169.2 \\
    3       & 55.6 & 98.7  & 72.4  & 162.1 \\
    4       & 56.6 & 99.2  & 72.9  & 158.9 \\
    \bottomrule
  \end{tabular}
\end{table}

\begin{table}
\centering
\small
\caption{\textbf{Ablation study on the impact of bounding box rescaling.} We report MPJPE (mm) on OOD datasets for models trained on H36M. "With Fix" refers to applying our rescaling method. The results show significant gains for SimpleBaseline and VideoPose3D on 3DPW, where the scale difference is largest.}
\label{tab:bbox_rescale_results}
% Set the inter-column separation
\setlength{\tabcolsep}{4pt}
% Use the tabular* environment set to the column width
\begin{tabular*}{\columnwidth}{@{\extracolsep{\fill}}l l r r r@{}}
\toprule
Model             & Dataset & No Rescaling & With Rescaling & Change \\
\midrule
SimpleBaseline    & 3DPW    & 164.5          & 147.4          & -10.0\%   \\
SimpleBaseline    & 3DHP    &  96.7          &  96.1          &  -0.6\%   \\
SimpleBaseline    & Fit3D   &  75.1          &  75.2          &  +0.1\%   \\
\midrule
VideoPose3D       & 3DPW    & 147.4          & 140.8          &  -4.5\%   \\
VideoPose3D       & 3DHP    &  89.9          &  89.8          &  -0.1\%   \\
VideoPose3D       & Fit3D   &  75.0          &  75.5          &  +0.7\%   \\
\midrule
PoseFormer        & 3DPW    & 126.7          & 131.8          &  +4.0\%   \\
PoseFormer        & 3DHP    &  84.1          &  86.1          &  +2.4\%   \\
PoseFormer        & Fit3D   &  70.3          &  71.0          &  +1.0\%   \\
\bottomrule
\end{tabular*}
\end{table}

\paragraph{Summary of Findings.}
The results of our ablation studies show that the SimpleBaseline~\cite{martinez2017simple} and VideoPose3D~\cite{pavllo2018video} models gain significantly from the bounding box rescaling, particularly on the 3DPW dataset. PoseFormer~\cite{Zheng2021PoseFormer} shows negligible benefit from this component. As discussed in the main text, MotionBERT~\cite{zhu2023motionbert} was omitted from this study as it already uses privileged focal-length and root-depth information in its camera-invariant codec, making our rescaling redundant.

\section*{Appendix D: Brittleness of Motion Cues}

This section details our preliminary analysis on the generalization of temporal lifting models, which motivated the design of AugLift. To investigate the common assumption that longer motion sequences improve generalization, we performed a controlled study using the VideoPose3D~\cite{pavllo2018video} architecture. To isolate the pure effect of motion dynamics from detection noise, all models were trained on ground-truth 2D keypoints from H36M~\cite{ionescu2014human3}. To account for viewpoint changes, we also applied perspective camera augmentations during training~\cite{gong2021poseaug}.

\begin{figure}[t]
    \centering
    \begin{minipage}{0.48\linewidth}
        \centering
        \includegraphics[width=\linewidth]{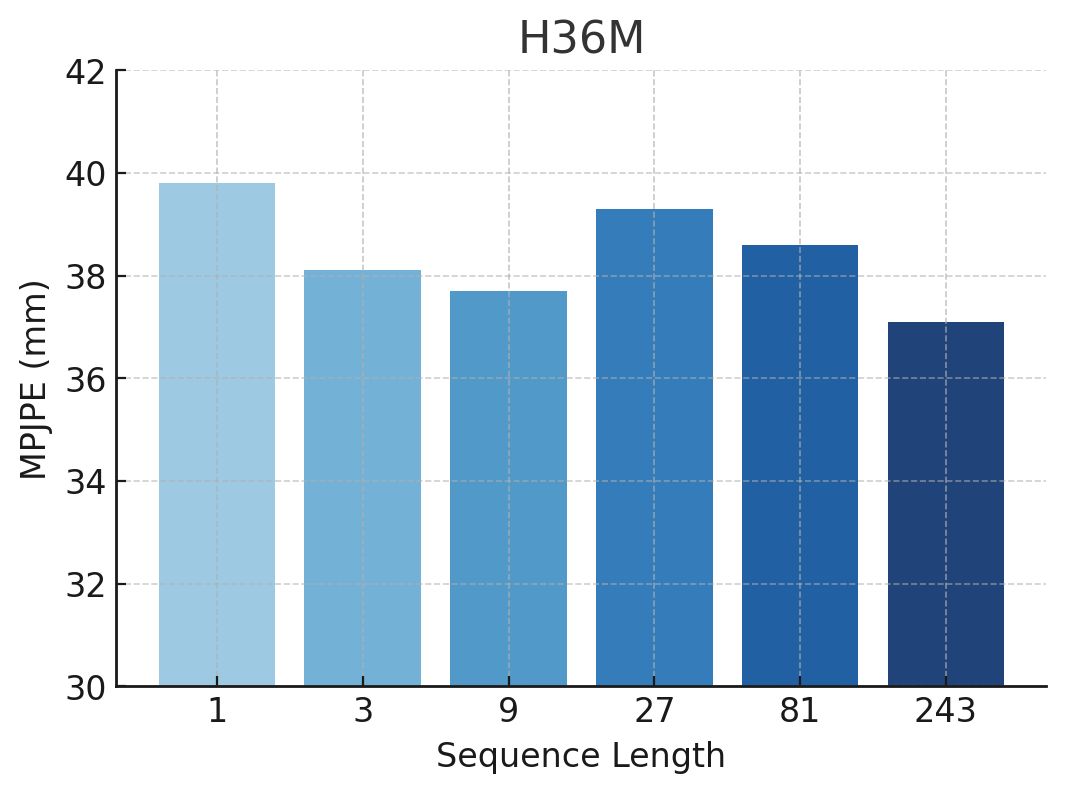}
    \end{minipage}%
    \begin{minipage}{0.48\linewidth}
        \centering
        \includegraphics[width=\linewidth]{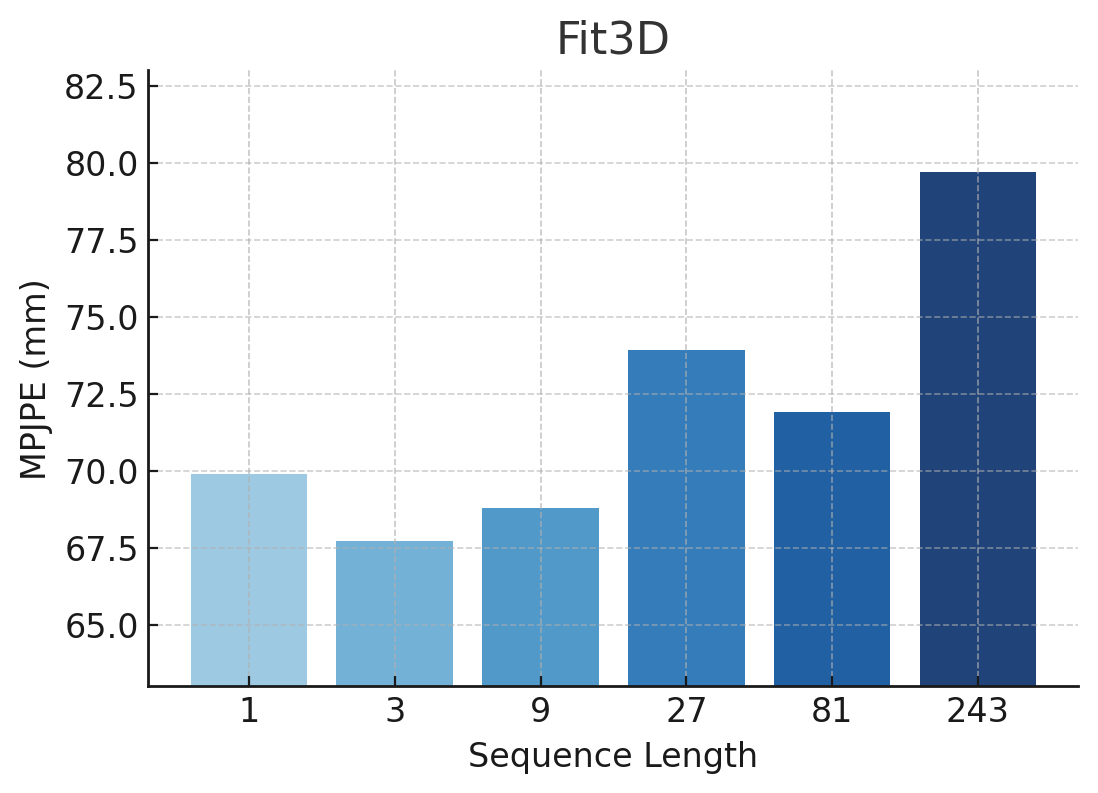}
    \end{minipage}
    \hfill
    \begin{minipage}{0.48\linewidth}
        \centering
        \includegraphics[width=\linewidth]{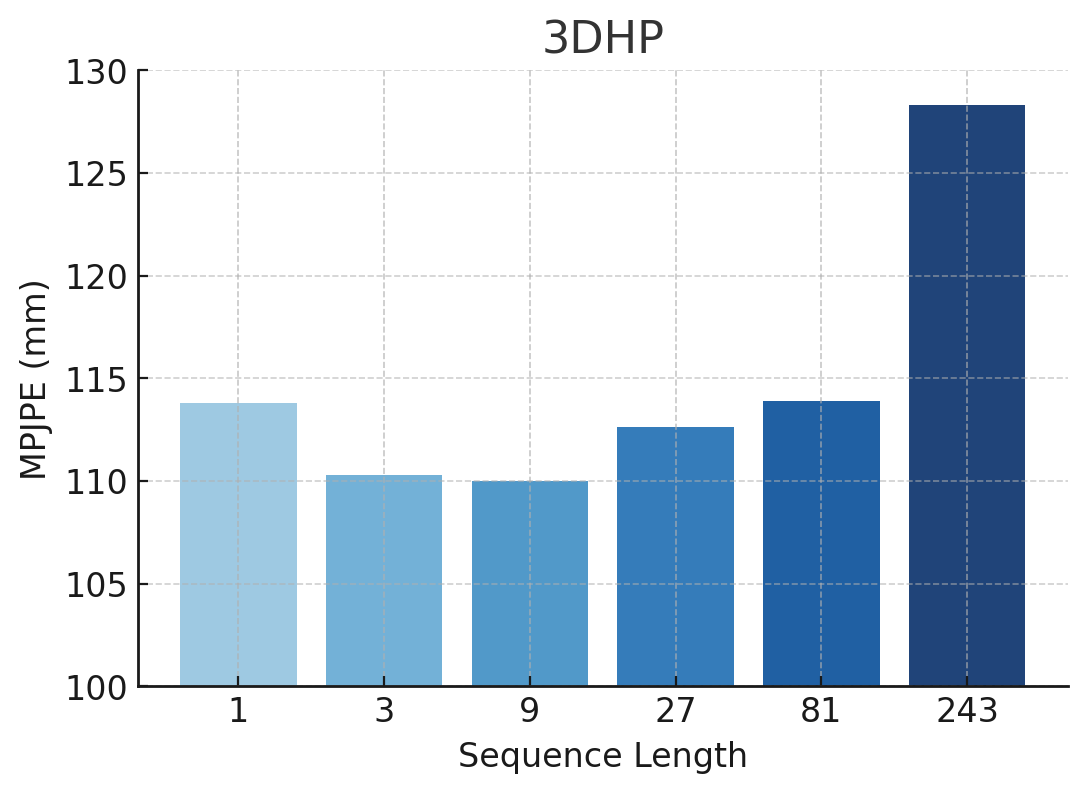}
    \end{minipage}
    \begin{minipage}{0.48\linewidth}
        \centering
        \includegraphics[width=\linewidth]{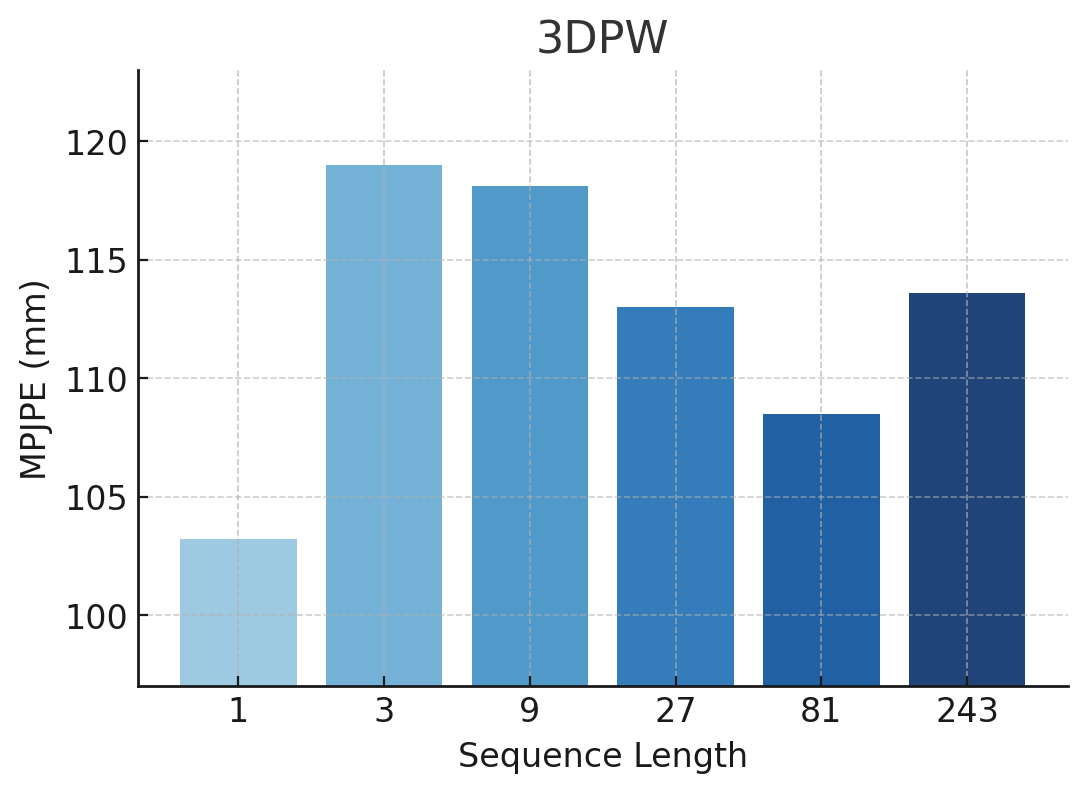}
    \end{minipage}
    \caption{\textbf{Longer sequences can harm cross-dataset generalization.}
    While in-distribution error (H36M, top-left) decreases with sequence length, out-of-distribution error on other datasets stays flat or increases.
    This analysis used a VideoPose3D model in the GT setting.}
    \label{fig:sequence_length_generalization_TCN}
\end{figure}

Our analysis reveals a surprising vulnerability of temporal models. As shown in Figure~\ref{fig:sequence_length_generalization_TCN}, while increasing the input sequence length consistently improves in-distribution performance on H36M, it fails to improve—and often degrades—cross-dataset performance on 3DHP, Fit3D, and 3DPW. This suggests models overfit to dataset-specific motion patterns. We tested this hypothesis by applying simple temporal transformations (e.g., reversing sequences or altering playback speeds) to create OOD motions from ID poses. Table~\ref{tab:part1_sequence_based_distortions_intra_dataset} confirms that models using longer sequences are more sensitive to these transformations, underscoring their brittleness to motion variations not seen during training.

\begin{table}%[ht]
\centering
\caption{\textbf{Overfitting to Motion Priors.} MPJPE on H36M validation using ground-truth 2D sequences. Simple temporal transformations (reversing, varying speed) significantly degrade performance for models using longer sequences, demonstrating overfitting to motion patterns.}
\label{tab:part1_sequence_based_distortions_intra_dataset}
\small
\begin{tabular}{lcccc}
    \toprule
    \textbf{Split} & \textbf{Seq 1} & \textbf{Seq 27} & \textbf{Seq 81} & \textbf{Seq 243} \\
    \midrule
    Train (Original)      & 25.1  & 16.6  & 15.7  & 13.4  \\
    Train (Reversed)      & 25.1  & 19.6  & 20.3  & 20.0  \\
    Train (Var. Speeds) & 25.2  & 30.4  & 32.9  & 32.1  \\
    \midrule
    Val (Original)        & 39.8  & 39.3  & 38.6  & 37.1  \\
    Val (Reversed)        & 39.8  & 41.3  & 40.9  & 39.5  \\
    Val (Var. Speeds) & 39.8  & 49.6  & 50.4  & 49.3  \\
    \bottomrule
\end{tabular}
\end{table}

\vspace{8pt}

\begin{figure}[!ht]
    \centering
    \includegraphics[width=1.1\columnwidth]{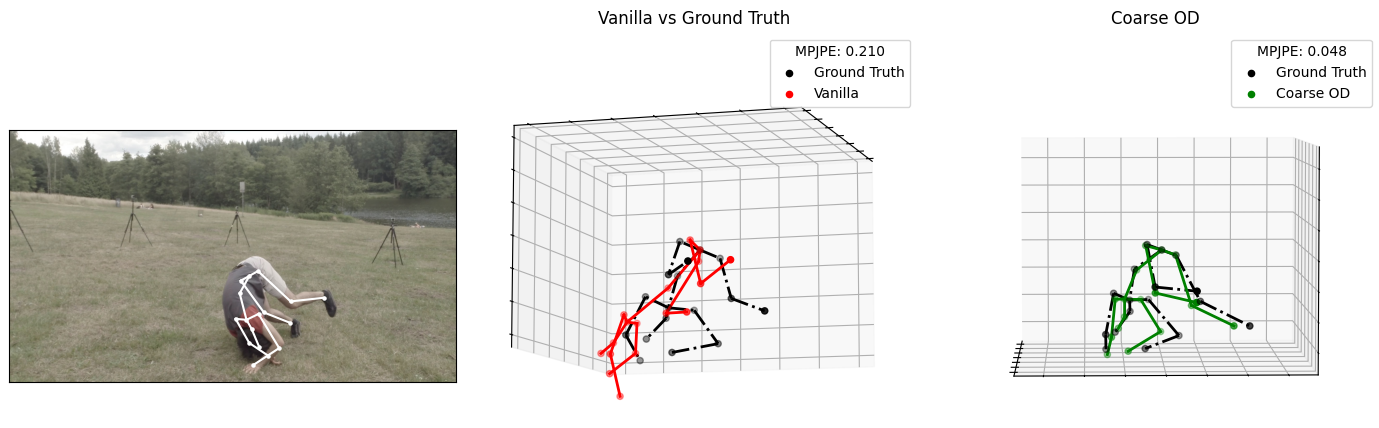}
    \includegraphics[width=1.1\columnwidth]{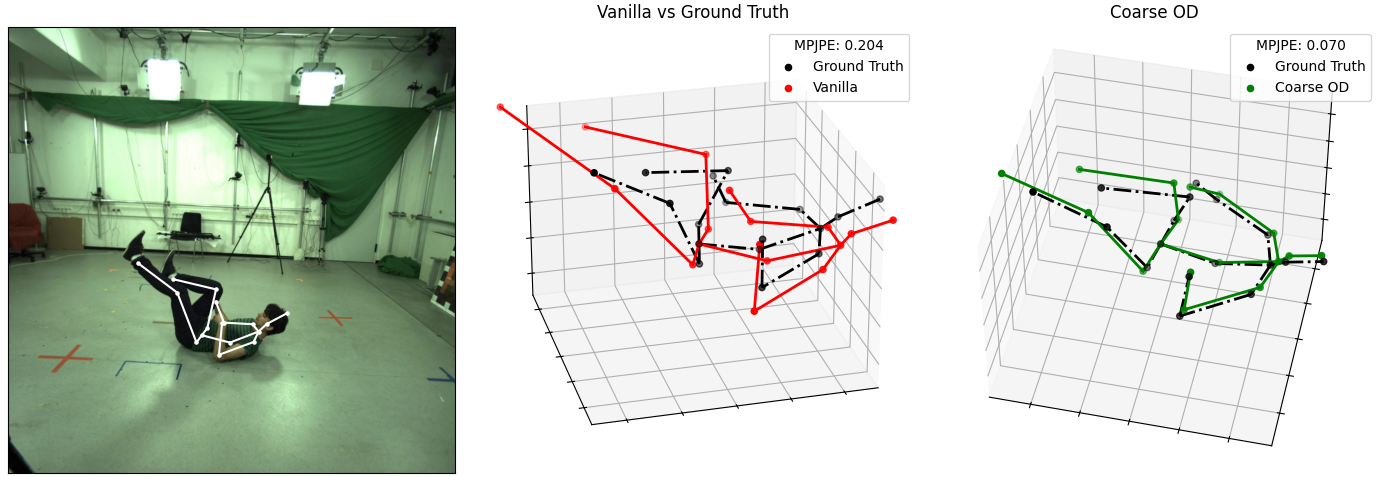}
    \includegraphics[width=1.1\columnwidth]{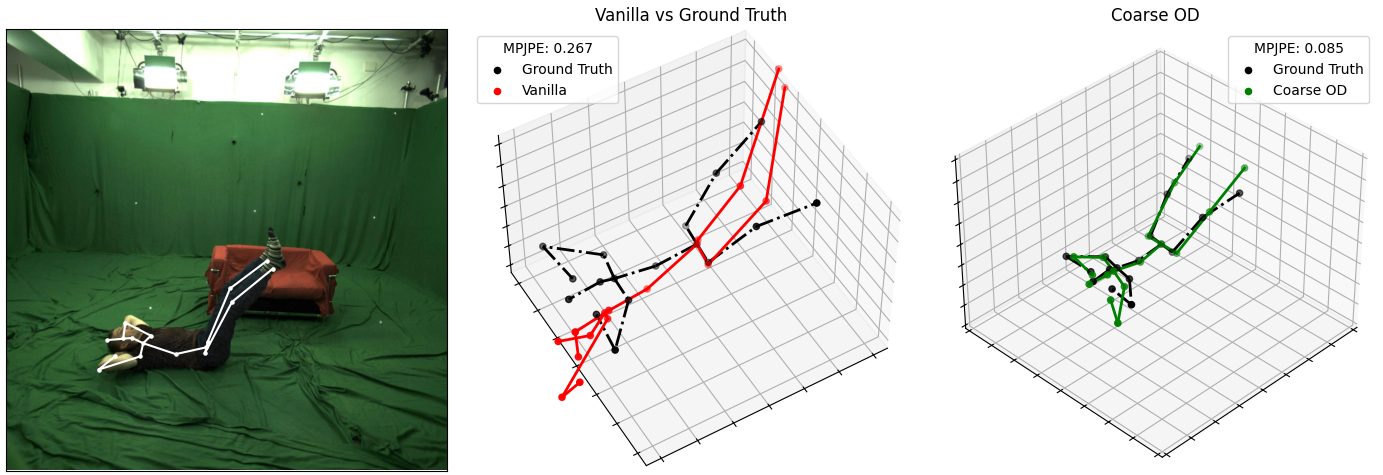}
    \caption{\textbf{Coarse ordinal depth (OD) greatly improves generalization by reducing depth ambiguity.}
    Each row compares the baseline lifting model (Vanilla) to the model augmented with oracle Coarse OD on 2D ground truth keypoints.}
    \label{fig:ordinal_depth_qualitative}
\end{figure}

\section*{Appendix E: Deeper Dive into Spatial Cues}
\label{sec:depth_analysis}

We investigated the potential of enriching the per-frame spatial input. We focused on ordinal depth (OD), which orders keypoints by relative depth (e.g., closer or farther from the camera) without estimating precise metric values. While prior work has used OD as a supervision signal~\cite{pavlakos2018ordinal}, its potential as a direct input to lifting models has been less explored. To quantify the maximum potential of this approach, we conducted an oracle study where we added a coarse OD signal to the 2D skeleton input. In this setup, keypoints are grouped into three categories—in front of, at, or behind the pelvis—using a fixed threshold on the ground-truth 3D data (e.g., 10cm). The compelling results of this study, which showed a significant reduction in generalization error, formed the primary motivation for AugLift.

\subsection*{E.1 ~~ Coarse Ordinal Depth experiments}
\label{OD_Results}
Recall that this coarse ordinal depth (OD) signal is provided by an oracle, offering a lightweight spatial cue to complement the 2D keypoints. We tested the impact of this signal by discretizing each of the 17 joints into depth bins of varying coarseness, ranging from 1 cm to 25 cm. The results in Table~\ref{tab:coarseness_impact} show that even at a coarse 25 cm granularity, the OD signal reduces generalization error on 3DHP by nearly 25\%, despite requiring only 2.9 bins per frame on average. This suggests that even a rough estimate of depth ordering—distinguishing which joints are in front of, in line with, or behind the pelvis—is highly informative for lifting models. Figure~\ref{fig:ordinal_depth_qualitative} provides a qualitative visualization of the significant improvements in the predicted 3D pose that result from leveraging these coarse OD cues.

\subsection*{E.2 ~~ Ablations on Coarseness of Signal}
We study the impact of the amount of coarseness to the ordinal depth bins, to cross dataset performance in Table \ref{tab:coarseness_impact}. While finer OD maps provide more depth detail, even a coarse three-bin OD (torso-depth, closer, farther) improves generalization, reducing MPJPE on 3DHP by 24.5\% compared to standard lifting. This suggests that even lightweight ordinal depth cues provide strong spatial constraints, helping models better infer depth without overfitting to dataset-specific motion.

\subsection*{E.3 ~~ Comparison with Prior Work}
In Table \ref{tab:coarse_od_generalization_vp3d}, we compare CoarseOD to perspective augmentations, and to Pavlakos 2018 \cite{pavlakos2016coarse}, demonstrating the effectiveness of adding depth cues and augmentation techniques in lifting-based 3D pose estimation. All models are trained on H36M and tested on 3DHP-INF, ensuring that performance is evaluated in a cross-dataset setting. The sequence length for TCN is set to 1 to isolate the impact of depth cues without reliance on temporal information. We report 3DPCK-150mm and AUC, as Pavlakos et al. do not provide MPJPE/P-MPJPE for the cross-dataset evaluation. Coarseness is set to 10cm, allowing for a balance between spatial granularity and robustness. The results show that CoarseOD + Perspective Aug achieves the best performance.

% \vspace{14pt}
\begin{table}%[ht!]
\centering
\caption{\textbf{Coarser ordinal depth maps still improve generalization.} We analyze the impact of ordinal depth (OD) coarseness on cross-dataset performance.}
\label{tab:coarseness_impact}
    \resizebox{\columnwidth}{!}{%
    \begin{tabular}{|l|c|c|c|}
        \hline
        \textbf{Coarseness} & \textbf{GT Bins (H36M)} & \textbf{GT Bins (3DHP)} & \textbf{MPJPE (3DHP)} \\
        \hline
        1 cm   & 13.7  & 13.8  & 52.6  \\  \hline
        10 cm  & 5.4   & 5.6   & 54.2  \\ \hline
        25 cm  & 2.9   & 2.9   & 61.2  \\
        \hline
    \end{tabular}%
    }
\end{table}

% \vspace{14pt}
\begin{table}%[!ht]
\centering
\caption{\textbf{Oracle CoarseOD significantly boosts cross-dataset performance.} CoarseOD improves generalization by 25\% or more, showing that adding depth cues and perspective augmentation enhances cross-dataset performance. Models trained on H36M and tested on 3DHP-INF benefit from these cues, with CoarseOD + Perspective Aug achieving the highest accuracy.}
% \niko{Also goes to show that depth cues and perspective augmentation can be complementary, motivating the broad applicability of our main analysis.}
\label{tab:coarse_od_generalization_vp3d}
\resizebox{\columnwidth}{!}{%
\begin{tabular}{lcccc}
\toprule
\textbf{Method} & \textbf{MPJPE} $\downarrow$ & \textbf{PMPJPE} $\downarrow$ & \textbf{3DPCK} $\uparrow$ & \textbf{AUC} $\uparrow$ \\
& (mm) & (mm) & (150mm, \%) &  \\
\midrule
Pavlakos 2018 & --- & --- & 71.9 & 35.3 \\
Vanilla & 94.4 & 63.1 & 81.4 & 48.8 \\
Perspective Aug & 81.1 & 57.4 & 86.8 & 53.9 \\
CoarseOD & 60.2 & 40.5 & 95.4 & 61.6 \\
\textbf{CoarseOD + Perspective Aug} & \textbf{53.1} & \textbf{36.9} & \textbf{97.2} & \textbf{65.5} \\
\bottomrule
\end{tabular}%
}
\end{table}

\section*{Appendix F: Inference Latency and Ablations}
% \niko{to organize this more}
\label{sec:other_results}

\subsection*{F.1 ~~ Inference Latency}
\label{sec:inference_constraints}
% \paragraph{Inference Latency.}
A key consideration is whether the generalization gains from AugLift justify its additional computational cost. We argue they do. The added latency is manageable, especially when compared to other components in the pipeline. For context, a common 2D detector like HRNet-W48 requires ~44\,ms for inference, and the MotionBERT lifter itself takes ~16.7\,ms. In contrast, using an efficient depth model like Depth Anything v2 (small variant) adds only ~13.3\,ms of latency, a cost that is further amortized in streaming applications where depth maps can be computed periodically and reused across frames. While a more thorough analysis is warranted, this modest computational overhead is a worthwhile trade-off for the substantial OOD performance improvement. Moreover, for a given latency budget, with AugLift, developers can now reduce the sequence length in favor of AugLift's rich spatial cues and achieve superior out-of-distribution performance. % Additionally, developers can now choose to reduce sequence length in favor of AugLift's rich spatial cues, achieving superior out-of-distribution performance within a given latency budget.

\subsection*{F.2 ~~ Ablations}
\paragraph{Impact of Bounding Box Rescaling.}
Our bounding box rescaling (Step 3 in Algorithm~1) is designed to normalize for variations in camera distance, a common challenge in OOD settings. Indeed, dataset statistics reveal significant differences in subject depth across benchmarks (Table~\ref{tab:dataset-stats}). Subjects in 3DPW, for example, are much closer to the camera (mean depth 3.29m) than in the H36M training set (5.67m), resulting in out-of-distribution bounding box sizes. While prior work like PoseDA~\cite{chai2023global} has addressed this, their approach requires access to test set summary statistics. In contrast, our method performs a per-instance normalization, using the on-the-fly depth estimates to adapt to each test example individually without privileged information. As shown in Table~\ref{tab:bbox-rescaling}, this provides significant gains on the 3DPW dataset where the domain gap is largest, reducing error by up to 14.3\%. Crucially, the gains from this rescaling are independent of whether the core AugLift cues are used, suggesting it is a broadly beneficial component in combating train-test domain gaps. We note, however, that it does not benefit all architectures; for instance, MotionBERT already incorporates a camera-invariance mechanism, making our rescaling redundant.

\vspace{6pt}
\begin{table}[t]
  \centering
  \caption{\textbf{Dataset statistics reveal a significant domain gap in subject depth.} The mean depth for 3DPW differs substantially from the primary training set, H36M.}
  \label{tab:dataset-stats}
  {%
    \begin{tabular}{l cc}
      \toprule
      Dataset  & Mean Depth & Depth (10\%--90\%) \\
      \midrule
      H36M     & 5.67m      & 4.58--6.53m      \\
      3DHP     & 5.05m      & 3.74--6.31m      \\
      Fit3D    & 5.00m      & 4.45--5.28m      \\
      3DPW     & 3.29m      & 2.20--4.50m      \\
      \bottomrule
    \end{tabular}%
  }
\end{table}

\begin{table}[t]
    \centering
    \small
    % \caption{\textbf{AugLift's sparse cues remain additive at longer sequences.}
    % Results are for PoseFormer (sequence length 9) trained on H36M and evaluated on H36M validation (Val, ID) and 3DHP (OOD).
    % RTMPose denotes image features from the 2D detector; AugLiftV2 is omitted for this sequence length.
    % }
    \caption{\textbf{AugLift's sparse cues remain additive in the presence of dense features at longer sequence lengths.}
Using PoseFormer with sequence length 9, we evaluate whether image features (from RTMPose) and depth–confidence cues (AugLift) provide complementary gains. }

% RTMPose features alone improve both ID and OOD accuracy, but AugLift provides larger generalization gains, and combining the two yields the best OOD performance. AugLiftV2 offers the strongest ID improvement in this regime, though its interaction with image features at longer temporal horizons remains an open direction.}

    \label{tab:feature_map_fusion_seq9}
    \setlength{\tabcolsep}{3.5pt}
    \renewcommand{\arraystretch}{0.95}
    \begin{tabular}{lcccc}
    \toprule
    Config              & Val & $\Delta_{\mathrm{ID}}$ (\%) & 3DHP & $\Delta_{\mathrm{OOD}}$ (\%) \\
    \midrule
    Base Model          & 52.2 & --    & 91.4 & --    \\
    Base Model + RTMPose& 50.9 & +2.5 & 88.1 & +3.6 \\
    % Base Model + DAV2   & 49.0 & +6.1 & 90.6 & +0.9 \\
    \midrule
    AugLift             & 50.5 & +3.3 & 84.9 & +7.1 \\
    AugLift + RTMPose   & 49.9 & +4.4 & \textbf{84.0} & +8.1 \\
    \midrule
    AugLiftV2     & \textbf{48.5} & +7.1 & 87.6 & +4.1 \\
    \bottomrule
    \end{tabular}
    \vspace{-5pt}
\end{table}

% \vspace{-6pt}
% \setlength{\intextsep}{-4pt} % vertical space above & below [h] floats
\begin{table}[h]
  \centering
  \caption{\textbf{Per-instance bounding box rescaling improves performance on the OOD 3DPW dataset.} We report MPJPE (mm) with and without our on-the-fly rescaling enabled. It is broadly beneficial, even in the absence of other AugLift components.}
  % \caption{\textbf{Per-instance bounding box rescaling improves performance on the OOD 3DPW dataset.} We report MPJPE (mm) with and without our on-the-fly rescaling enabled.}
  \label{tab:bbox-rescaling}
  \vspace{-5pt}
  \begin{tabular}{ll ccr}
    \toprule
    \multicolumn{2}{l}{} & \multicolumn{2}{c}{Rescaling} & \\
    \cmidrule(lr){3-4}
    Architecture & Variant & No & Yes & $\Delta$ (\%) \\
    \midrule
    SimpleBaseline  & Baseline & 182.5 & 156.4 & 14.3\% \\
    SimpleBaseline  & AugLift & 164.5 & 147.4 & 10.0\% \\
    \midrule
    VideoPose3D     & Baseline & 168.3 & 161.4 & 4.0\%  \\
    VideoPose3D     & AugLift & 147.4 & 140.8 & 4.5\%  \\
    \bottomrule
  \end{tabular}
\end{table}

\begin{figure}[t]
  \centering
  \includegraphics[width=0.5\textwidth]{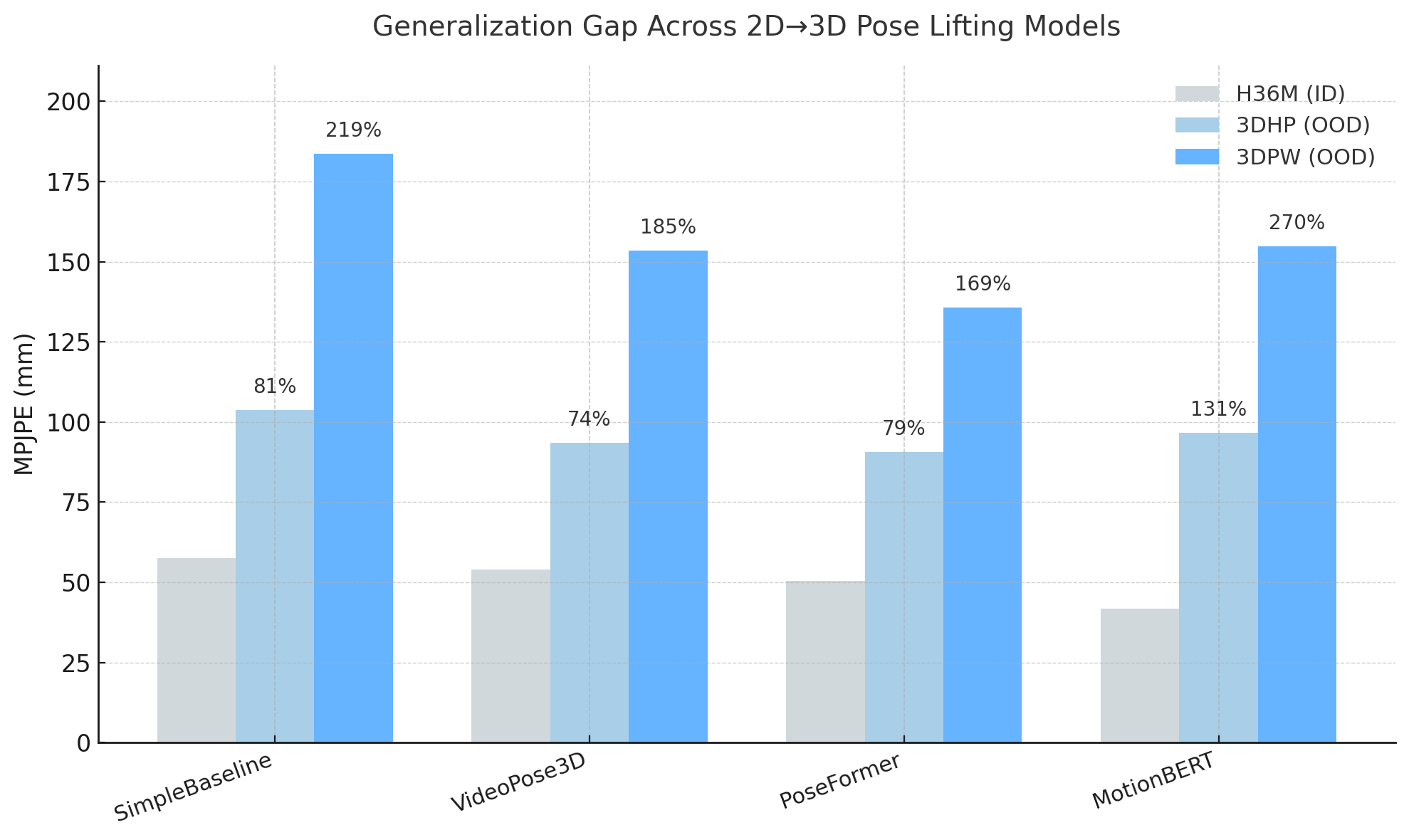}
  \caption{%
    \textbf{Modern lifting models generalize poorly across datasets.} We plot the in-distribution (ID) MPJPE on H3.6M against out-of-distribution (OOD) performance on 3DHP and 3DPW for four popular architectures. Percentages show the OOD error inflation, with even state-of-the-art models exhibiting gaps of over 150\%. See Table 2 and Section~4.3 for detailed results.%
  }
  \label{fig:generalization_gap}
\end{figure}

% \begin{table}[t]
%     \centering
%     \small % Makes the font size for the caption and table smaller
%     \caption{\textbf{AugLift's sparse cues are complementary to image and dense depth features.} AugLift's sparse cues are complementary to image and dense depth features.
% We analyze fusing our sparse AugLift (UADD) with dense image features (+RTMPose) and explore rich feature fusion in the depth map space, an underexplored area, with our AugLiftV2.}
%     \label{tab:feature_map_fusion_combined_copy}
%     \setlength{\tabcolsep}{3.5pt} % Adjust column spacing
%     \renewcommand{\arraystretch}{0.95} % Adjust row height
%     \begin{tabular}{lcccc}
%     \toprule
%     Config & Val & $\Delta_{\mathrm{ID}}$ (\%) & 3DHP & $\Delta_{\mathrm{OOD}}$ \\
%     \midrule
%     Base Model & 54.3 & -- & 94.5 & -- \\
%     Base Model + RTMPose & 51.1 & +5.9 & 88.1 & +6.8 \\
%     \midrule
%     AugLift & 51.7 & +4.8 & 87.9 & +7.0 \\
%     AugLift + RTMPose & 50.3 & +7.4 & 84.6 & +10.5 \\
%     \midrule
%     AugLiftV2 & 47.9 & +11.8 & 87.1 & +7.8 \\
%     AugLiftV2 + RTMPose & 47.9 & +11.8 & 85.9 & +9.1 \\
%     \bottomrule
%     \end{tabular}

% \end{table}

\paragraph{Impact of Sequence Length.}
We now analyze how AugLift's benefits vary with the amount of temporal information available to the model. In this study, we use the default settings from Section~4.1 (MotionBERT architecture trained on H36M) and evaluate it across a range of sequence lengths. For each sequence length, we report the in-distribution (ID) improvement on H36M and the mean out-of-distribution (OOD) improvement averaged over the three other test sets (3DHP, Fit3D, and 3DPW).

The results, summarized in Table~\ref{tab:improvements_fixed}, reveal a key trend: while the in-distribution gains from AugLift taper as more temporal context is added, the out-of-distribution gains remain substantial. For single-frame models (Seq-Len 1), AugLift reduces OOD error by a significant \textbf{11.7\%}. Even for very long sequences (Seq-Len 243), where the baseline model is already very strong, AugLift still delivers a robust \textbf{8.3\%} OOD error reduction.
% This strongly supports our hypothesis that the rich, per-frame spatial cues provided by AugLift provide a more generalizable signal than potentially brittle motion priors (discussed in Section~\ref{ssec:prelim_analyses}), fortifying the model to OOD shifts.
Furthermore, the final column ($\Delta_{\text{Depth}}$) shows that adding only the depth cue to a model that already has confidence and temporal context continues to provide a large OOD boost (e.g., 6.0\% at Seq-Len 243). This reinforces that the depth signal provides unique geometric information that is not captured by motion alone.

\begin{table}[t]
\vspace{-4pt}                % lift table toward preceding text
\setlength{\tabcolsep}{3pt}  % reduce horizontal padding
\renewcommand{\arraystretch}{0.85} % reduce vertical padding
\centering
\caption{\textbf{AugLift's OOD benefits are substantial across all sequence lengths.} We show the percentage improvement (\%) for MotionBERT trained on H36M. $\Delta_{\text{AugLift}}$ is the gain of the full model over the baseline. $\Delta_{\text{Depth}}$ shows the marginal gain from adding the depth cue to a model that already uses confidence.}
\begin{tabular}{l l r r}
\toprule
Seq-Len & Split & $\Delta_{\text{AugLift}}$ & $\Delta_{\text{Depth}}$ \\
\midrule
1   & ID  &  5.0\%  & 6.7\%  \\
1   & OOD &  11.7\% & 9.2\%  \\
\midrule
9   & ID  &  4.3\%  & 2.8\%  \\
9   & OOD &  7.6\%  & 4.7\%  \\
\midrule
27  & ID  &  2.7\%  & 1.9\%  \\
27  & OOD &  7.5\%  & 5.8\%  \\
\midrule
243 & ID  &  1.4\%  & 1.8\%  \\
243 & OOD &  8.3\%  & 6.0\%  \\
\bottomrule
\end{tabular}
\label{tab:improvements_fixed}
\vspace{0pt}
\end{table}

\section*{Appendix G: Other Results Feature Fusion}
% \niko{to organize this more}
\label{sec:other_results_feature_fusion}

% \niko{2DO: cleanup the writing here}
\begin{figure}[t]
  \centering
  \includegraphics[width=0.5\textwidth]{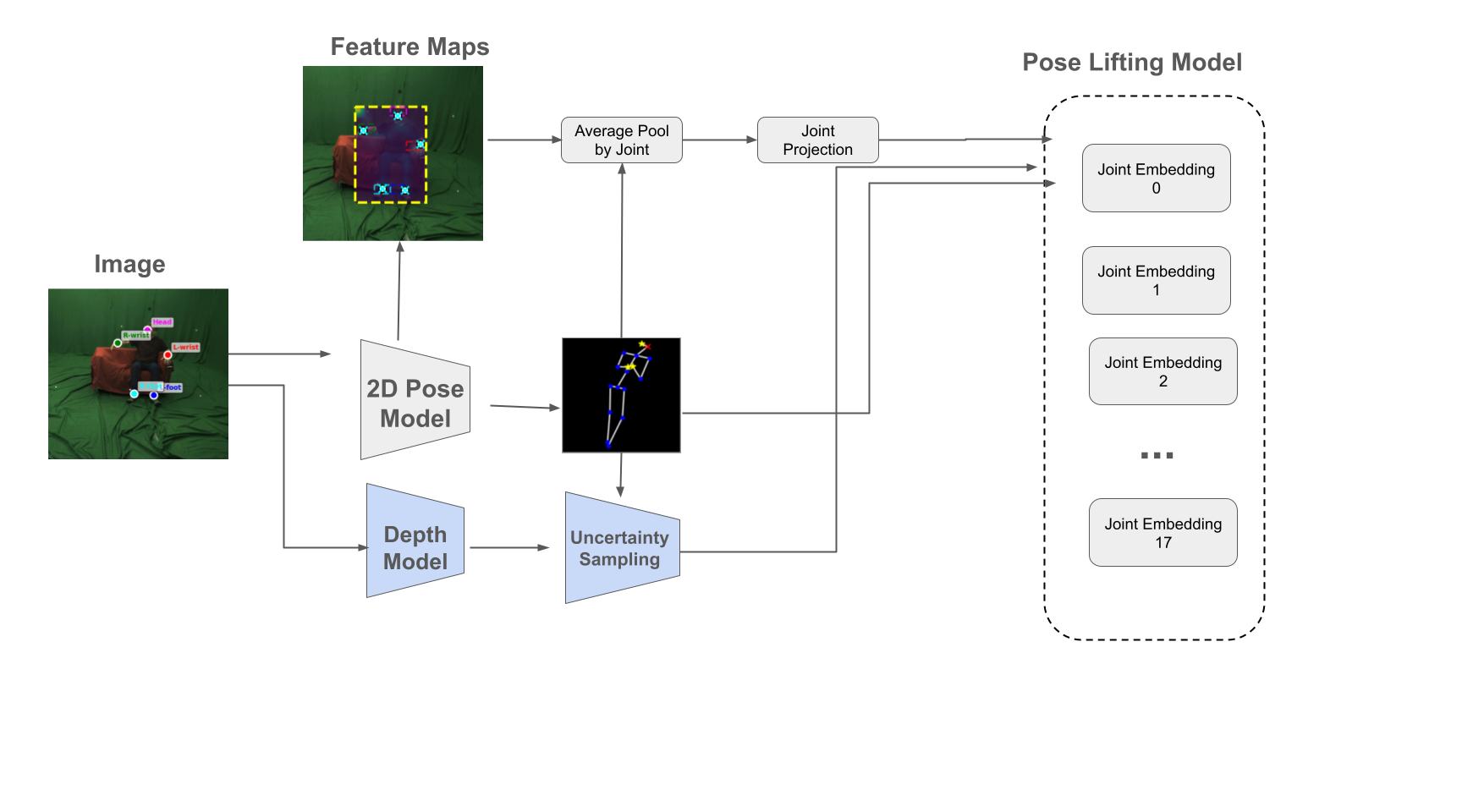}
  % \caption{System diagram illustrating feature fusion.}
  \vspace{-4pt}
  \caption{\textbf{Feature fusion pipeline used in our lifting experiments.}
For each detected 2D keypoint, we align the corresponding coarse feature map from RTMPose or DAV2 and bilinearly sample the four nearest spatial patches to obtain a localized dense feature vector.
% These sampled features are average-pooled, linearly projected to a compact 16-D embedding, and concatenated with either the raw $(x,y)$ input or the full AugLift descriptor before entering the PoseFormer encoder.
This design provides a lightweight, modular mechanism for injecting appearance or depth cues at the joint level and preserves the efficiency of standard lifting architectures.}
  \label{fig:feature_fusion_arch}
\end{figure}

Our dense-feature experiments also highlight two considerations. First, although AugLiftV2 introduces learned depth-image features beyond the handcrafted $(c,d)$ cues of UADD, the added computation remains lightweight: sampling coarse feature maps and projecting them to 16 dimensions introduces only a modest, practically negligible increase in inference latency. Second, combining confidence–depth statistics with learned depth-image embeddings is, to our knowledge, a novel formulation.

We extend our lifting models with dense feature maps extracted from the final backbone layers of both RTMPose and Depth Anything v2 (DAV2). For RTMPose, we use its final coarse feature map of shape $(1024, 9, 12)$ and bilinearly interpolate the four nearest spatial patches surrounding each 2D keypoint to obtain a 1024-D vector. For DAV2, we use its final depth-backbone feature map with spatial dimensions $(37 \times 37)$; we first reduce the channel dimension using a lightweight 1D convolution to 256 channels for efficiency, then sample the four nearest spatial features at each keypoint using the same interpolation strategy.

All sampled feature vectors are passed through a learned linear projection layer that maps them to a compact 16-D embedding. These projected features are then concatenated with either the base $(x,y)$ keypoint input or the full AugLift descriptor $(x,y,c,d,d_{\min},d_{\max})$ before entering the PoseFormer spatial encoder.
% This design keeps feature fusion lightweight, maintains compatibility with standard lifting architectures, and corresponds exactly to the experimental settings used in Section~5.

Figure~\ref{fig:feature_fusion_arch} illustrates the feature-fusion pipeline used in our experiments. For each detected 2D keypoint, we align the corresponding coarse feature map from RTMPose or DAV2 and bilinearly sample the four nearest spatial patches, yielding a dense feature vector centered at the lifting keypoint. These vectors are then passed through a lightweight linear projection to a 16-D embedding and concatenated with the keypoint token (either $(x,y)$ or the full AugLift descriptor) before entering the PoseFormer encoder. This design provides a clean and modular way to inject localized appearance or depth information into lifting, while keeping the architecture simple and computationally efficient.

Table~\ref{tab:feature_map_fusion_seq9} shows that at a longer sequence length (Seq.\,Len\,9), AugLift's sparse depth–confidence cues remain complementary to dense image features. While RTMPose features alone provide modest improvements, AugLift consistently yields stronger generalization gains, and combining AugLift with image features produces the best OOD performance. AugLiftV2 achieves the largest ID gains in this regime, though its interaction with image features at longer sequence lengths remains an open question. These preliminary results suggest that even when temporal context is available, sparse depth cues continue to provide additive benefits over dense feature fusion.

\section*{Appendix 7: AugLift+ Domain Generalization Details}
\label{sec:appendix7}

\subsection*{7.H ~~ PoseAug Framework and Analytical SMPL Mapping}
\label{sec:poseaug_smpl_details}

This subsection provides full details on the PoseAug augmentation framework and our analytical SMPL integration, summarized in Section~5 of the main paper.

\paragraph{PoseAug augmentation framework.}
PoseAug~\cite{gong2021poseaug} applies three sequential differentiable transformations during training: \textbf{(1) RT (rotation-translation)}: random 3D viewpoint transformations via rotation matrices and camera translation; \textbf{(2) BA (bone angle)}: an MLP predicts joint angle residuals that perturb skeletal directions while preserving bone connectivity; \textbf{(3) BL (bone length)}: an MLP predicts bone length scaling ratios to simulate skeleton size variation across individuals. These augmentations operate directly in 3D pose space before projection to 2D, creating diverse viewpoint and anatomical variations that encourage learning of domain-invariant features.

\paragraph{Analytical SMPL integration.}
To generate depth-augmented training data, we employ an analytical BA/BL$\to$SMPL mapping that converts PoseAug's augmented 3D skeletons into SMPL body model parameters, achieving roughly two orders of magnitude speedup over iterative optimization. For BA augmentation, we need to find the SMPL joint rotation that reproduces each augmented bone direction. At branching joints---pelvis (3 children: left hip, right hip, spine) and thorax (3 children: neck, left shoulder, right shoulder)---a single rotation must simultaneously align multiple outgoing bones, so we solve for the optimal rotation via SVD-based Procrustes alignment of the child bone directions. At serial-chain joints (limbs, spine segments) with only one child, we use the closed-form Rodrigues rotation between the original and target bone direction. Each resulting rotation is converted to SMPL's 23$\times$3 axis-angle \texttt{body\_pose} representation, with coordinate conversion between camera and SMPL space via $F = \mathrm{diag}(1,-1,-1)$.

\paragraph{BL: rest-pose bone scaling.}
For BL augmentation, we analytically scale the SMPL rest-pose skeleton without optimizing shape parameters ($\beta$). For each SMPL bone $(p, j)$ with a corresponding H36M bone, we compute the scale factor from the augmented and original H36M bone lengths:
\begin{equation}
s_{(p,j)} = \frac{\|\mathbf{b}_{\mathrm{aug}}\|}{\|\mathbf{b}_{\mathrm{orig}}\|}
\end{equation}
We then update the rest-pose joint positions in topological (parent-first) order:
\begin{equation}
J_{\mathrm{new}}[j] = J_{\mathrm{new}}[\mathrm{parent}(j)] + s_{(p,j)} \cdot \bigl(J_{\mathrm{rest}}[j] - J_{\mathrm{rest}}[\mathrm{parent}(j)]\bigr)
\end{equation}
Of the 23 non-root SMPL bones, 18 have H36M correspondences; the remaining 5 (feet, hands, head top) retain their original lengths ($s=1$). This direct geometric scaling avoids any iterative optimization of shape parameters.

\paragraph{Rendering pipeline.}
We render each augmented SMPL mesh using the PyTorch3D differentiable rasterizer at $128 \times 128$ resolution (the subject spans ${\sim}50$\,px, sufficient for 17-joint depth sampling). We use the HardPhongShader ($5.7\times$ faster than SoftPhongShader) on a SMPL mesh decimated from 13{,}776 to ${\sim}$3{,}000 faces via quadric decimation, with focal length $f = 5000 \cdot r / 256$ where $r$ is the render resolution. The rendered RGB image is passed through Depth Anything V2 (ViT-S) to produce a depth map in the same domain as real-image depth predictions. Per-joint depth values are extracted via bilinear interpolation at the projected 2D keypoint locations on this depth map, with min-pooling for edge tolerance. The resulting root-relative depth channel, clipped to $\pm 2$\,m, is concatenated with the 2D keypoints $(x,y)$ to form the 3-channel XYD input to the lifting network.

\subsection*{7.I ~~ AugLift+ Evaluation Details}
\label{sec:augliftplus_eval_details}

For the 3DHP results reported in Section~5, we evaluate on the full 3DHP test set (test-all), following the protocol of AdaptPose, which provides a more comprehensive assessment than the commonly used test\_valid subset.

\paragraph{3DPW evaluation protocols.}
The 3DPW results in Table~4 use the PoseBench~\cite{manzur2025posebench3d} protocol (35{,}515 frames, SMPL-24 joints). However, the standard convention from the HMR literature uses the \texttt{J\_regressor} matrix to regress SMPL meshes into H36M-compatible joint format, enabling direct comparison of 3D pose accuracy without skeleton format differences. Table~\ref{tab:jreg_comparison} compares both protocols across all evaluated models. The SMPL-24 skeleton has substantially shorter hip-to-neck distance (281\,mm vs.\ 482\,mm for J\_regressor), resulting in systematically higher absolute MPJPE values in the PoseBench protocol.

\begin{table}[h]
\centering
\small
\caption{3DPW cross-dataset evaluation comparing J\_regressor and PoseBench protocols (MPJPE $\downarrow$, mm). All models trained on H36M with GT 2D keypoints.}
\label{tab:jreg_comparison}
\begin{tabular}{l|cc}
\hline
Method & j\_reg & PB \\
\hline
\multicolumn{3}{l}{\textit{Baselines}} \\
VPose3D (temporal) & 120.3 & 121.4 \\
Simple Baseline (no aug) & -- & 137.5 \\
\hline
\multicolumn{3}{l}{\textit{Published methods}} \\
PoseAug MLP & 74.4 & 121.2 \\
DAF-DG feedback MLP & 115.4 & 127.0 \\
AdaptPose VPose3D (3DHP) & 87.7 & 112.7 \\
AdaptPose VPose3D (3DPW) & 85.3 & 104.6 \\
\hline
\multicolumn{3}{l}{\textit{Ours}} \\
AugLift + PoseAug (XYD) & -- & 104.3 \\
AugLift + PoseAug (XYD+dscale2d) & \textbf{63.3} & \textbf{92.6} \\
\hline
\end{tabular}
\end{table}

The ``--'' entries indicate configurations not evaluated under that protocol; specifically, the XYD model without dscale2d was evaluated only on PoseBench. AdaptPose trains separate models targeting each test domain.

\subsection*{7.J ~~ DAF-DG Reproduction}
\label{sec:dafdg_reproduction}

We attempted to reproduce DAF-DG~\cite{peng2024daf} (CVPR 2024), which reports state-of-the-art cross-dataset performance (63.1\,mm on 3DHP, 106.6\,mm on 3DPW). The released code contains multiple crash bugs (undefined variables) and logic bugs that impact the meta-optimization component central to the method. After fixing all crash bugs and exploring multiple variants to restore functional meta-optimization, all configurations either diverged or degraded steadily over training. Our best stable result achieves 80.5\,mm on 3DHP. An independently filed GitHub issue (\url{https://github.com/davidpengucf/DAF-DG/issues/1}) reports that ``test results were even higher than PoseAug.''

% Bibliography is handled by the main document — do not include here.

\end{document}